\documentclass[journal]{IEEEtran}
\usepackage{amsfonts}
\usepackage{amsmath}
\usepackage{booktabs}
\usepackage{multirow}
\usepackage{algorithm}
\usepackage{algorithmic}
\usepackage{array}
\usepackage{makecell}
\usepackage{bm}
\usepackage{graphicx}
\usepackage{epsfig}
\usepackage{subfigure}
\usepackage{cite}
\usepackage{color}
\usepackage[para, online, flushleft] {threeparttable}
\usepackage{amssymb}
\usepackage{url} 
\usepackage{verbatim}
\usepackage{xspace}
\usepackage{diagbox}

\newcolumntype{L}[1]{>{\raggedright\arraybackslash}m{#1}}
\newcolumntype{C}[1]{>{\centering\arraybackslash}m{#1}}
\newcolumntype{R}[1]{>{\raggedleft\arraybackslash}m{#1}}
\newcommand{\RB}[1]{\textbf{\textcolor[rgb]{1,0,0}{#1}}}
\newcommand{\GB}[1]{\textbf{\textcolor[RGB]{0, 100, 0}{#1}}}
\newcommand{\BB}[1]{\textbf{\textcolor[rgb]{0,0,1}{#1}}}

\makeatletter
\DeclareRobustCommand\onedot{\futurelet\@let@token\@onedot}
\def\@onedot{\ifx\@let@token.\else.\null\fi\xspace}

\def\eg{\emph{e.g}\onedot} 
\def\ie{\emph{i.e}\onedot}

\def\wrt{w.r.t\onedot} 
\def\etal{\emph{et al}\onedot}
\makeatother

\ifCLASSINFOpdf
\else
\fi
\begin{document}
%
\title{Self-explanatory Deep Salient Object Detection}
%
%
%

\author{Huaxin~Xiao,
        Jiashi~Feng,
		Yunchao~Wei,
		Maojun Zhang
\thanks{Huaxin Xiao and Maojun Zhang are with the college of Information System and Management, National University of Defense Technology, China (e-mail: xiaohuaxin@nudt.edu.cn; mjzhang@nudt.edu.cn).}
\thanks{Jiashi Feng and Yunchao Wei are with Department of Electrical and Computer Engineering, National University of Singapore, Singapore (e-mail: elefjia@nus.edu.sg; eleweiyv@nus.edu.sg).}
}

%
%

\markboth{Journal of \LaTeX\ Class Files}%
{Xiao \MakeLowercase{\textit{et al.}}}
%



\maketitle

\begin{abstract}
Salient object detection has seen remarkable progress driven by deep learning techniques.  However, most of deep learning based salient object detection methods are black-box in nature and lacking in interpretability. This paper proposes the first self-explanatory saliency detection network that explicitly exploits low- and high-level features for salient object detection. We demonstrate that such supportive clues not only significantly enhances performance of salient object detection but also gives better justified detection results. More specifically, we develop a multi-stage \textit{saliency encoder} to extract multi-scale features which contain both low- and high-level saliency context. Dense short- and long-range connections are introduced to reuse these features iteratively. Benefiting from the direct access to low- and high-level features, the proposed saliency encoder can not only model the object context but also preserve the boundary. Furthermore, a \textit{self-explanatory generator} is proposed to interpret how the proposed saliency encoder or other deep saliency models making decisions. The generator simulates the absence of interesting features by preventing these features from contributing to the saliency classifier and estimates the corresponding saliency prediction without these features. A comparison function, \textit{saliency explanation}, is defined to measure the prediction changes between deep saliency models and corresponding generator. Through visualizing the differences, we can interpret the capability of different deep neural networks based saliency detection models and demonstrate that our proposed model indeed uses more reasonable structure for salient object detection. Extensive experiments on five popular benchmark datasets and the visualized saliency explanation demonstrate that the proposed method provides new state-of-the-art.
\end{abstract}


%
\IEEEpeerreviewmaketitle

\section{Introduction\label{sec1}}
%
%
%
%
\IEEEPARstart{S}{a}lient object detection aims to localize and segment the most visually attractive objects from a given image,  simulating the visual attention process in human vision systems. Segmenting out conspicuous objects extensively facilitates numerous computer vision tasks by filtering irrelevant distracting information, such as for object tracking \cite{mahadevan2009saliency}, image retrieval \cite{gao20123}, weakly-supervised object segmentation \cite{wei2017object, oh2017exploiting}, to name a few.

\begin{figure}[t]
	\centering
	\subfigure[Source]{\includegraphics[width=1.63cm]{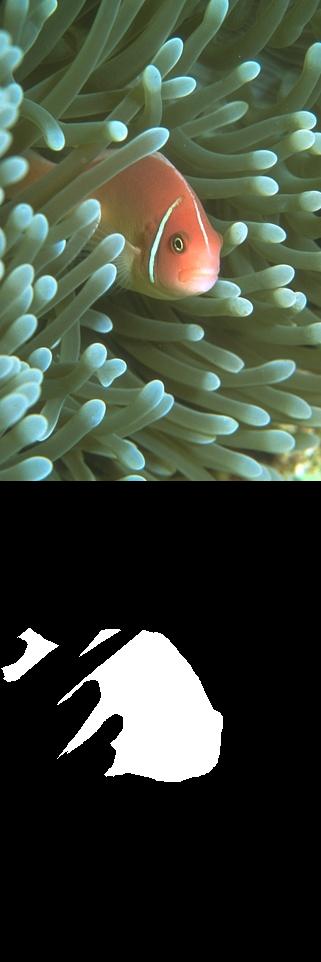}\label{f1a}}
	\subfigure[DHS~\cite{liu2016dhsnet}]{\includegraphics[width=1.63cm]{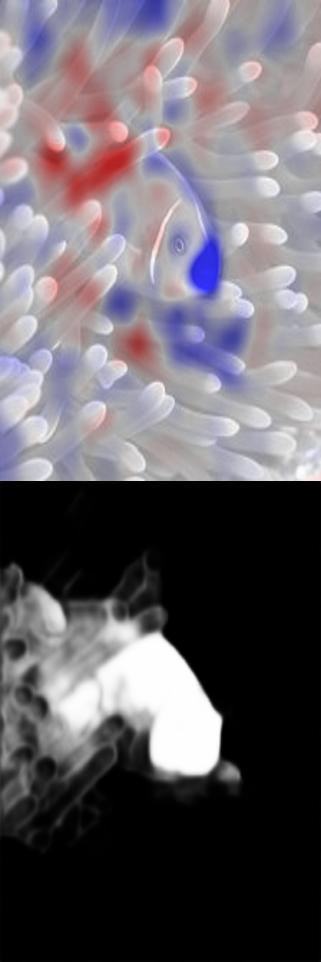}\label{f1c}}
	\subfigure[DCL~\cite{li2016deep}]{\includegraphics[width=1.63cm]{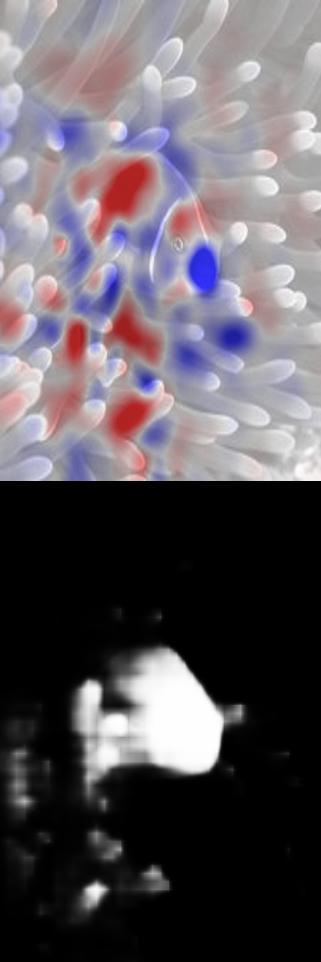}\label{f1b}}
	\subfigure[DSS~\cite{hou2016deeply}]{\includegraphics[width=1.63cm]{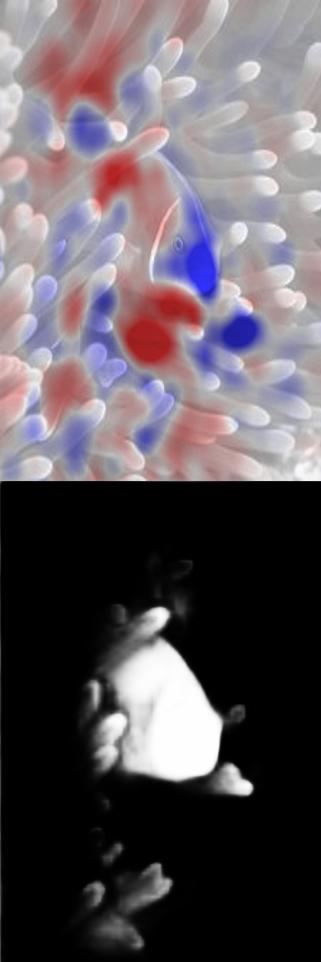}\label{f1d}}
	\subfigure[SENet]{\includegraphics[width=1.63cm]{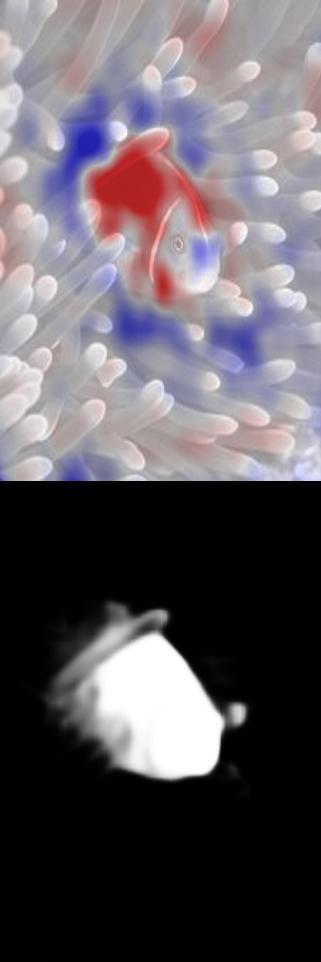}\label{f1e}}
	\caption{Visual comparison of saliency detection results and saliency explanation by different models. In the first row, we visualize the derived saliency explanation that interprets how the saliency prediction made by DHS~\cite{liu2016dhsnet}, DCL~\cite{li2016deep}, DSS~\cite{hou2016deeply}, and the proposed SENet. The red areas mean these regions provide positive support for salient object detection while the blue ones represent that the existing of this region could suppress and distract saliency detection. The white color areas mean no significant sensitivity for salient object detection. The second row shows the saliency detection results without any post-processing. The proposed SENet extracts clues from object for saliency detection and thus provides more accurate and integral detection results. Best viewed in color. \label{fig1}}
\end{figure}

Recently, remarkable progress for salient object detection has been made with the development and deployment of deep neural networks (DNNs)~\cite{wang2015deep, zhao2015saliency, lee2016deep, li2016deep, tang2016saliency, kuen2016recurrent, liu2016dhsnet, wang2016saliency, hou2016deeply, li2017instance, luo2017non, hu2017deep, li2015visual, li2016visualtip} of various architectures. Among them, some DNN based models~\cite{wang2015deep, zhao2015saliency, li2015visual} simply determine whether a local region is salient or not through a single classifier on top of its hierarchically extracted representation. The performance of such models is severely limited by the narrow ``receptive field'' over the raw input image due to ignoring important context clues. To remedy deficiency of the narrow perspective by only applying a network locally, another global networks in ~\cite{wang2015deep, zhao2015saliency, li2015visual} are introduced to provide global object context. Besides relying on regional prediction, some pixel-level dense prediction networks were developed and adopted. For instance, the fully convolutional networks (FCNs)~\cite{long2015fully} can produce dense prediction by associating  each pixel  with a classifier. One of the advantages offered by such dense prediction networks is to preserve rich spatial and local details that are recognized to be crucial for saliency detection \cite{liu2016dhsnet, luo2017non}. 
Upon this kind of models, diverse refinements, \eg, introducing recurrent units~\cite{kuen2016recurrent, wang2016saliency} and superpixel-based filtration~\cite{li2016deepsaliency, li2016deep, hu2017deep}, are proposed to better exploit fine details of the salient object.
Due to the various challenges in salient object detection, \eg, low contrast between the salient object and the background, difficult situations in the non-salient regions and multiple salient objects (will be discussed in Section~\ref{sec_viscomp}), how to properly integrate the local and global perspective of saliency context is still crucial for improving performance. To better exploit multi-level clues without incurring  additional network engineering efforts or expensive refinement operations, this paper introduces a novel end-to-end architecture where features from multiple levels are directly and recursively integrated to provide more powerful support for detecting salient objects.


Besides developing a novel deep model for salient object detection, we observe that all of existing DNN based  models are lacking in interpretability because of their ``black-box'' nature.  Enhancing  transparency of deep models would offer reasonable explanations on their predictions and in turn benefit  these models  in following  two aspects. First, revealing  the internal mechanism of deep saliency detection models can tell which part of an image is most influential (either positively or negatively) for prediction making. Such explanation is helpful to understand the effectiveness of different architectures in  saliency prediction. Secondly, understandable and transparent property of DNNs will encourage saliency detection model to exploit really useful cues for prediction  and  inspire novel network architectures. Hence, we argue that  {a desired deep saliency detection model should not only detect the salient object accurately but also provide self-explanations on its predictions}.



To develop a salient object detection model with both superior performance and clear interpretability, we propose a Self-Explanatory saliency detection Network (SENet) in this work. The  SENet consists of two essential components: a \textit{saliency encoder} and a \textit{self-explanatory generator}. The saliency encoder  predicts saliency scores for each pixel from the input image, or in other words it encodes a raw input image into a saliency map. Within the encoder, we propose to explicitly widen access of the saliency classifier to multi-level clues by efficiently reusing multi-scale features through novel dense short- and long-range connections. These multi-range dense connections facilitate the encoder to make saliency predictions by exploiting all computed feature maps, which is beneficial for modeling the saliency context and preserving object boundary.  Besides, all features are explicitly connected to the classifier via dense connections. Such architecture can improve the model training process as  the gradients from the final classifier are directly propagated backwards to  features at multiple levels\textemdash which can effectively alleviate the notorious  vanishing gradient issue~\cite{huang2017densely}. Benefited from this novel architecture of the encoder,  our proposed SENet can detect the salient region with high confidence and produce a complete and sharp salient region, as illustrated by  the example given in the second row of Figure \ref{fig1}.

To enhance model interpretability, a self-explanatory generator is introduced for SENet that infers an influence distribution over the learned representations from the saliency encoder. As a probabilistic model, the generator  estimates the probability of encoder predictions with absence of some input features. A comparison function, \textit{saliency explanation}, is defined to track the prediction changes. In this way, the saliency explanation can clearly present the positive or negative effect of these absent features over the saliency predictions. More importantly, after going through all input features, we can integrate these results to explain why particular saliency prediction is made for the entire features (\eg, an image). 

We provide example saliency detection results from different models along with explanation on their decision to better illustrate the idea. As shown in the first row of Figure \ref{fig1}, we apply the self-explanatory generator to provide explanation for three state-of-the-arts and the SENet. The interesting features in the generator are the raw input image. The red areas indicate the regions that provide positive support for salient object detection, and the blue ones represent that such regions could suppress and distract the performance of saliency detection. The white color areas mean no significant sensitivity for saliency detection. From the visualized explanation, one can observe that the proposed SENet correctly relies on the \textit{ fish} (salient object) for saliency prediction and regards the surroundings of the salient object as negative factors. This means leaving out the local area around the fish can improve saliency prediction. This is understandable intuitively as leaving the area around the object out would make the object more salient. For the compared models in Figure \ref{fig1}, the positive areas (which influence the model decision most significantly) spread among salient and non-salient region that lead to false positive detection results in saliency maps. These visualization results well explain why those models give inferior saliency detection performance compared with ours. 

In short, the main contributions of this paper can be summarized as follows:
\begin{itemize}
	\item We develop a saliency detection model that is self-explanatory. Different from existing deep saliency detection models that are generally ``black-box'', our proposed model is able to provide reasonable explanation for saliency detection. Results on benchmark datasets clearly demonstrate that the proposed model provides a new state-of-the-art.
\end{itemize}

\begin{itemize}
	\item  We propose a new end-to-end deep saliency encoder with dense connections. It enriches the access of classifier to the encoded features by dense short- and long-range connections. These dense connections introduce multi-level features reusing and implicit deeply supervision which makes the network effective and efficiency. 
\end{itemize}

\begin{itemize}
	\item We proposed a generator to explain saliency predictions. The generator estimates the encoder prediction with the absence of some interesting features. Trough visualizing the differences between the predictions of generator and encoder, we can offer interpretable evidence for understanding the underlying basis of saliency predictions.
\end{itemize}

The rest parts of this paper are organized as follows. Section~\ref{sec2} reviews existing DNN based saliency detection approaches and the methods to interpret DNNs. Section~\ref{sec3} presents the details of the proposed SENet. Extensive experimental results on 5 popular benchmarks, the network interpretation and comparisons with state-of-the-art are presented in Section~\ref{sec4}. And Section~\ref{sec5} concludes this paper.

\section{Related Work}\label{sec2}
\subsection{DNN based Salient Object Detection}
Over the past years, DNNs have become powerful tools on various computer vision tasks and achieved state-of-the-art performance. Here, we focus on reviewing the related works that employ DNNs to detect  salient objects. 

One of popular schemes for DNN based saliency detection is to combine multi-networks for multi-scale saliency context. Different context networks can supplement each other and provide various information. Wang~\etal~\cite{wang2015deep} proposed a local network on cropped image patches for pixel-level saliency detection. To ensure label consistency, another network was trained to predict the saliency score of each object proposal from a global perspective. In~\cite{zhao2015saliency}, two DNNs were jointly trained and integrated to capture the global and local context. Li and Yu~\cite{li2015visual} aggregated three types of DNNs features from only salient object region, immediate neighboring region of salient object and the full image excluding the salient object respectively. Lee~\etal~\cite{lee2016deep} encoded the hand-crafted features by multiple convolutional units as supplement to the high-level DNNs features. Several fully connected layers were adopted to concatenate these features and classify the saliency region. In~\cite{li2016deep}, a segment-wise spatial pooling branch was proposed to refine the boundaries of salient object which was inferred by a multi-scale FCNs branch. Similarly, Tang and Wu~\cite{tang2016saliency} generated the segments in region-level network adaptively. A fusion network was trained to combine the pixel-level and region-level saliency map.

Another strategy for DNN based saliency detection is to refine the saliency results in stages. 
To obtain a entire object perception capability, Li~\etal~\cite{li2016deepsaliency} performed a multi-task learning scheme in conjunction with the task of semantic segmentation. Then the saliency map was refined by graph Laplacian regularized nonlinear regression to generate fine-grained boundaries. Kuen~\etal~\cite{kuen2016recurrent} exploited the spatial transformer and recurrent network units to iteratively attend to image region and refine it progressively. Liu and Han~\cite{liu2016dhsnet} proposed a hierarchical recurrent convolutional unit to refine the details of saliency map step by step. Wang~\etal~\cite{wang2016saliency} fed forward image and previous stage saliency map to the FCNs recurrently. The recurrent architecture of~\cite{wang2016saliency} made the network learn to correct the previous errors. In~\cite{kim2016shape}, region proposals were adopted to learn the boundaries of salient object. Then the coarse and regional saliency maps were merged and refined by image specific low-to-mid level information. Hu~\etal~\cite{hu2017deep} incorporated DNNs with level set function for learning more accurate boundaries and compact saliency. Then, the level set map was refined by a superpixel-based guided filter layer to propagate saliency.

Recently, a simplified and efficient DNN~\cite{luo2017non} was proposed to combines local and global information through a multi-resolution grid structure and a boundary penalty loss was implemented to ensure spacial consistency. Hou~\etal~\cite{hou2016deeply} exploited deeply supervision~\cite{lee2015deeply} with stage-wise short connections on multi-scale feature maps. A final saliency map was inferred by merging the middle side outputs while throwing away the deepest and the shallowest side outputs. Li~\etal~\cite{li2017instance} further developed salient object detection into instances level. A multi-scale shared saliency network was proposed to generate high-quality saliency masks and salient object contours. Based on the detected object contours, \cite{li2017instance}~generated the salient object proposals by a MAP-based subset optimization method to obtain the salient instances.

In this paper, we attempt to unify the multi-stages features by dense refinement. Compared to the multi-network scheme~\cite{wang2015deep, zhao2015saliency, li2015visual, tang2016saliency}, the proposed method need no extra network to model the salient object context or capture the boundary of  object. The refinement operation is completed by the dense and long connections which are more effective and efficient than the recurrent units \cite{kuen2016recurrent, wang2016saliency} or the regional filtration~\cite{li2016deepsaliency, li2016deep, hu2017deep}.

\subsection{ Explanation on DNN Models}
With the rapid progress of DNNs, the characteristic of non-transparency can be a barrier to further understand and improve DNNs. Hence, enhancing interpretabiity of deep learning models has emerged as an urgent requirement~\cite{zintgraf2017visualizing}. Generally, there are two approaches to understand DNNs: forward propagation based and backpropagation based \cite{shrikumar2017learning}.


For the forward propagation based approaches, they leave interesting features out and feed forward the rest parts of the features to the DNNs. The change of DNNs prediction explains the response of DNNs over the interesting features. Zeiler and Fergus \cite{zeiler2014visualizing} directly masked out parts of an image by a gray square, and monitoring the activation of intermediate layers or the output of the classifier. Based on the work of \cite{robnik2008explaining}, Zintgraf~\etal~\cite{zintgraf2017visualizing} adopted a more strict rule to leave out the interesting features. The absence of interesting features was formulated as Bayesian inference which was more reasonable than the direct gray patch \cite{zeiler2014visualizing}.

The backpropagation based approaches propagate the gradient from classifier towards interesting neuron or input image and visualize the activation to explain the DNNs. Based on the gradient of the certain class score \wrt the input image, Simonyan~\etal~\cite{simonyan2013deep} provided two types of explanation: produce an image that maximizes the interesting class prediction or compute an activation map for the interesting class. During the backpropagation of~\cite{simonyan2013deep}, the gradient of the ReLUs would be zero if the input to the ReLU is negative in the phase of forward propagation. To overcome the discontinued gradients of~\cite{simonyan2013deep}, Springenberg~\etal\cite{springenberg2014striving} proposed guided backpropagation to adds an additional guidance signal from the higher layers to usual backpropagation. Moreover, Shrikumar~\etal~\cite{shrikumar2017learning} proposed to compare the activation
of a neuron with its reference state to avoid the discontinuities in the gradient and highlight inputs that contribute positively or negatively to the output.

All above-mentioned methods focus on interpreting a single classifier for image classification. In this paper, we attempt to explain the behavior of pixel-wise dense classifiers. To this end, we extend the work of~\cite{zintgraf2017visualizing} to analyze the influence of interesting features over the dense classifiers and provide interpretation for the dense prediction, \ie, saliency detection. We fuse multiple explanation maps instead of the single category used by~\cite{zintgraf2017visualizing} and propose a comparison function, \ie saliency explanation, to measure the prediction changes of deep saliency models.

\begin{figure*}[t]
	\begin{center}
		\includegraphics[width=1.0\linewidth]{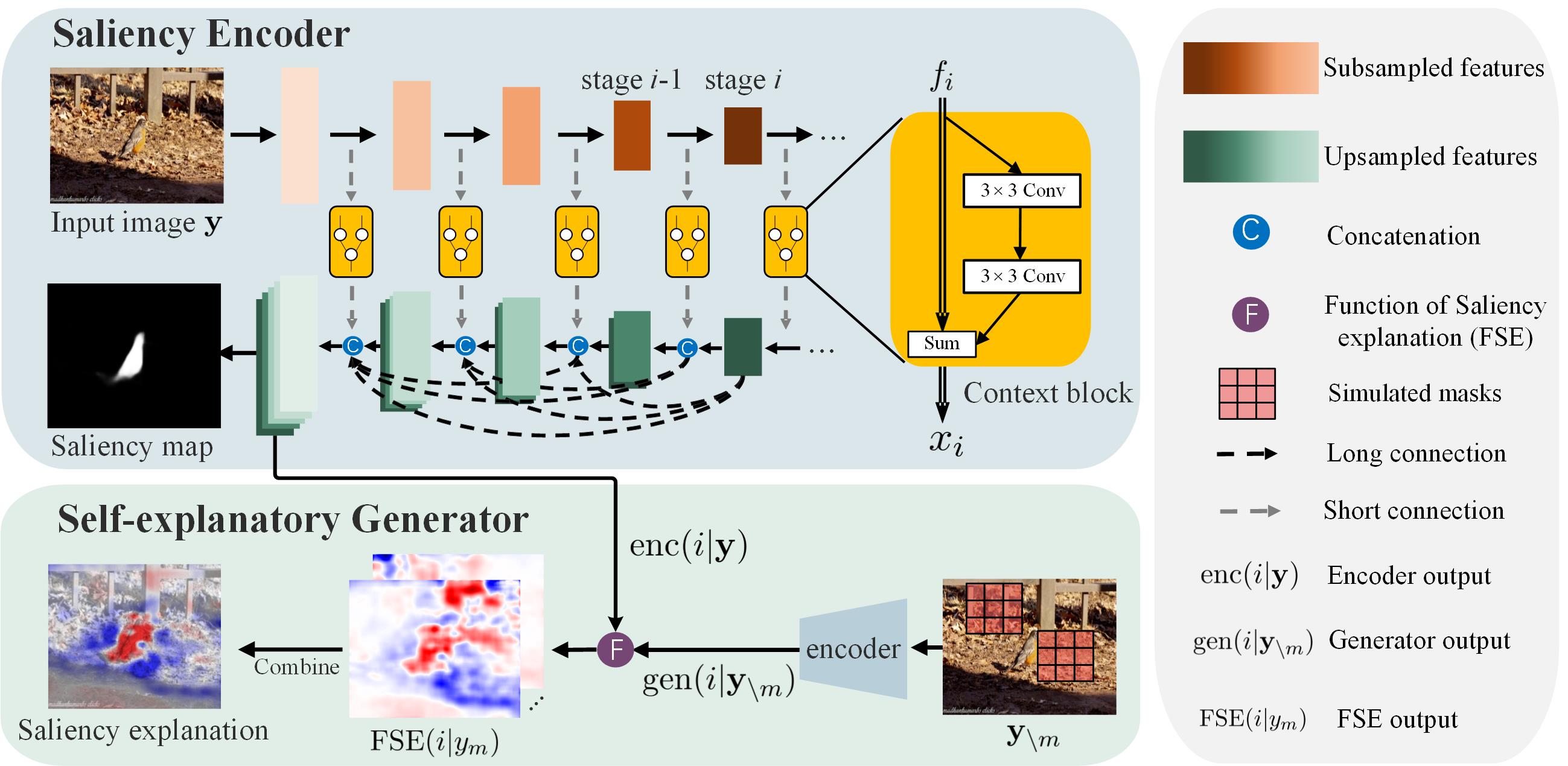}
	\end{center}
	\caption{
		Overall architecture of SENet. The saliency encoder employs a fully convolutional architecture to map the input image to a pixel-level saliency map. First, the saliency encoder passes the input image through a pre-trained image classification network and extracts features at multiple levels. A specifically designed context block is introduced to encode these features into saliency-diagnosis context features. Then, the encoded features are upsampled and concatenated to lower-level features iteratively by the dense short- and long-range connections. The self-explanatory generator is proposed to interpret the saliency prediction made by the encoder. For a given image, the generator purposively makes some parts of the image absent by preventing those parts from contributing to the encoder and tracks changes of the saliency prediction. After all parts of the image are examined, the discovered changes of the encoder are integrated to form the saliency explanation map. Best viewed in color.}
	\label{network}
\end{figure*}

\section{Proposed Self-explanatory Network}\label{sec3}

We start with introducing the overall architecture of the proposed SENet. We then proceed to provide details of the two novel and critical components, \ie, the encoder for saliency prediction and the generator for explanation, one by one.

\subsection{Overall Architecture}
The proposed SENet model consists of two novel components: the saliency  encoder for saliency detection and the self-explanatory generator that identifies explanation for supporting saliency detection. The overall architecture is shown in Figure~\ref{network}. The saliency encoder encodes an input image to a pixel-level saliency map where each element indicates  the probability of the corresponding pixel belonging to a salient object. The encoder employs a fully convolutional architecture to make the dense prediction and it works as follows. First, the saliency encoder passes  the input image through a pre-trained image classification network and extracts  features at multiple levels. A specifically designed context block is introduced to encode these features into saliency-diagnosis context features. Such encoded features are then upsampled and concatenated with  lower-level features recursively via newly proposed dense short- and long-range connections until the resolution of the features equals to the input image. These dense connections can ensure the saliency classifier fully exploit relevant context for making saliency prediction with enhanced accuracy.

The self-explanatory generator in Figure~\ref{network} is proposed to interpret the saliency prediction made by the encoder. For a given image, the generator purposively masks some parts of the image  by preventing those parts from contributing to the encoder and tracks changes of the saliency prediction. In this way, the supportive evidence used from the raw input image can be effectively identified. After all parts of the image are examined, the discovered changes of the encoder are integrated to form the saliency explanation map (as shown in the bottom panel of Figure~\ref{network}). Through visualizing the saliency explanation, the generator explicitly identifies the positive and negative factors of the given image that influence the saliency predictions. The positive factors mean the regions in the given image are heavily relied by the saliency models to make decision while the negative factors convey a message that removing these regions can bring a more accurate saliency result. That is, the presence of the negative regions may degrade the performance of saliency model. Moreover, the generator is compatible with other DNN based saliency models which can be adopted as a metric to compare the capability of different saliency models.


\subsection{Saliency Encoder}
The motivation behind the saliency encoder in SENet is to ensure the dense saliency classifier has access to both global and local perspective over the input image simultaneously when making saliency predictions. Both global and local information are crucial for saliency detection~\cite{zhao2015saliency, li2015visual} while a systematic method to comprehensively exploit these information is still absent. Therefore, we propose to integrate the information within the saliency detection models at multiple processing levels. A simple yet effective structure, called \emph{dense connections}, is adopted  to enrich the connection among intermediate features for fully utilizing the valuable information. In particular, two types of dense connections, \ie the short- and long-range connections, are introduced  to merge features at the same scale and transport features to other scales respectively. These dense connections avoid resorting to expensive ensemble of multiple networks  for different features or contexts~\cite{wang2015deep, zhao2015saliency, li2015visual, tang2016saliency} and also effectively get rid of additional refinement stages, \eg the superpixel-based filtration~\cite{li2016deepsaliency, li2016deep, hu2017deep} and the recurrent refinement~\cite{kuen2016recurrent, wang2016saliency}. When the saliency classifier makes decisions, the dense structure directly provides the classifier with various and necessary information instead of separating the decisions process of classifier into multi-networks or multi-stages. This end-to-end structure is thus more effective and efficient for salient object detection.

We now explain  architectural details of the saliency encoder component. We call all the subsampling operations involved between two adjacent feature maps  as forming one \emph{stage}. As shown in the top panel of  Figure \ref{network}, at the stage $i{-}1$, the saliency encoder in SENet receives and concatenates the features from all the higher stages $i, i+1, \cdots, L$ into a new feature: 
\begin{equation}\label{eq3}
x_{i-1}\leftarrow \left[x_{i-1}, H^{\mathrm{up}}_i(x_i), H^{\mathrm{up}}_{i+1}(x_{i+1}), \cdots, H^{\mathrm{up}}_{L}(x_L)\right],
\end{equation}
where $x_i$ is the feature map from the stage $i$, $H^{\mathrm{up}}_i(\cdot)$ is an upsampling transformation and $L$ is the total number of  stages in the encoder.  

These dense connections benefit the proposed encoder in the following desired aspects: (1) the saliency context can be reused at different  stages which helps the encoder effectively  collect information from both global and local perspectives and improve  saliency predictions; (2) the features at each stage have direct access to the gradients from the top loss layer  which can be regarded as implicit deep supervision \cite{lee2015deeply} and facilitates  learning features for saliency detection; (3) the dense connections can explicitly widen  flow of the gradient and make  model training easier~\cite{huang2017densely}.

The dense connections receive features from a pre-trained DNN model  for image classification. To make the extracted features more suitable for salient object detection, we fine-tune outputs of the classification network through a \emph{context block}. As demonstrated in the top panel of Figure \ref{network}, before the dense connections, the proposed saliency encoder  employs a residual-like block to augment the features. Concretely, after each  stage $i$, the feature maps $f_i$ are passed through a context block and give the following output  $x_i$:
\begin{equation}\label{eq1}
x_i = G_i\left(f_i;  W_i \right) + f_i, 
\end{equation}
where  $ W_i $ is the weight parameter of  the  context block at stage $i$ and $G_i(\cdot;\cdot)$ is a non-linear transformation over the input feature, as detailed below. 

 \begin{figure}[t]
 	\centering
 	\subfigure[]{\includegraphics[width=2.1cm]{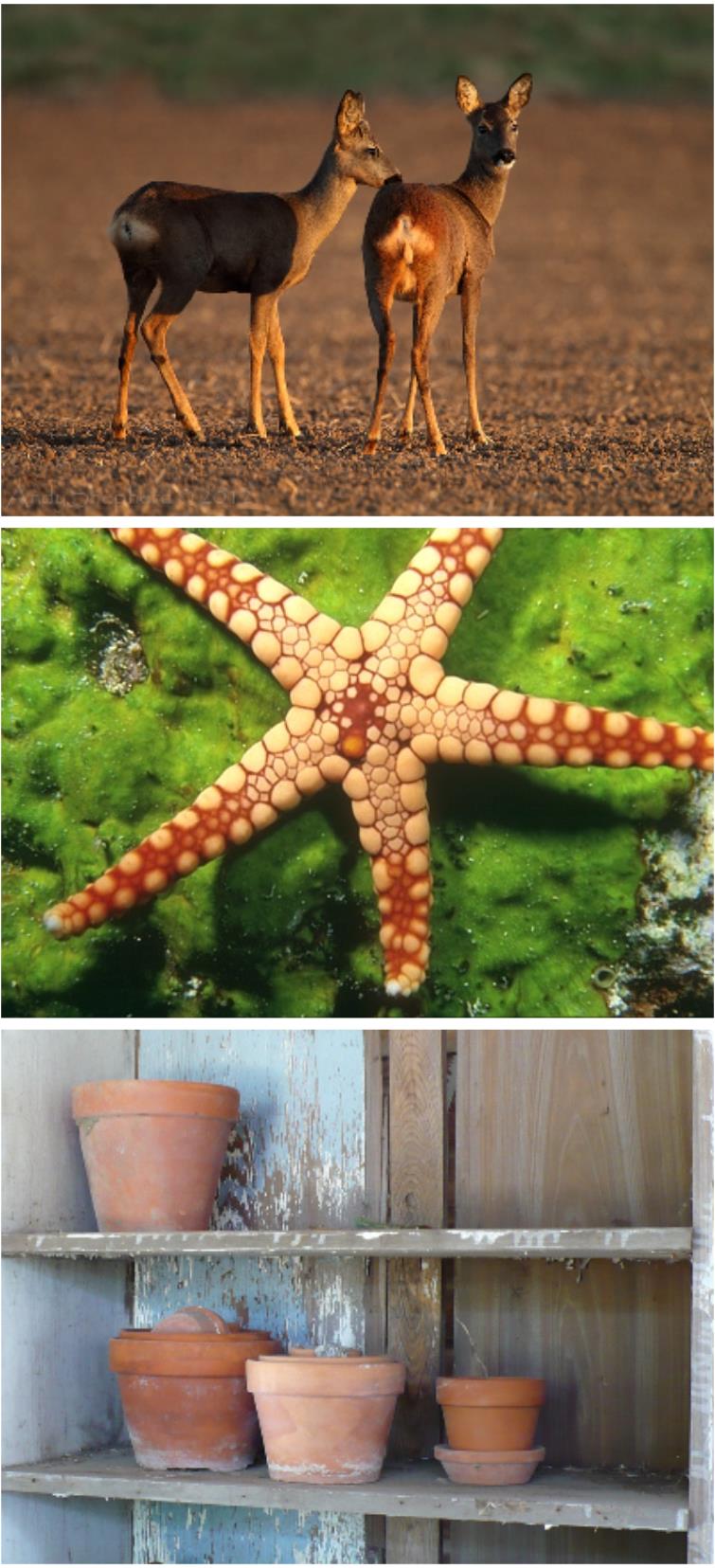}}
 	\subfigure[]{\includegraphics[width=2.1cm]{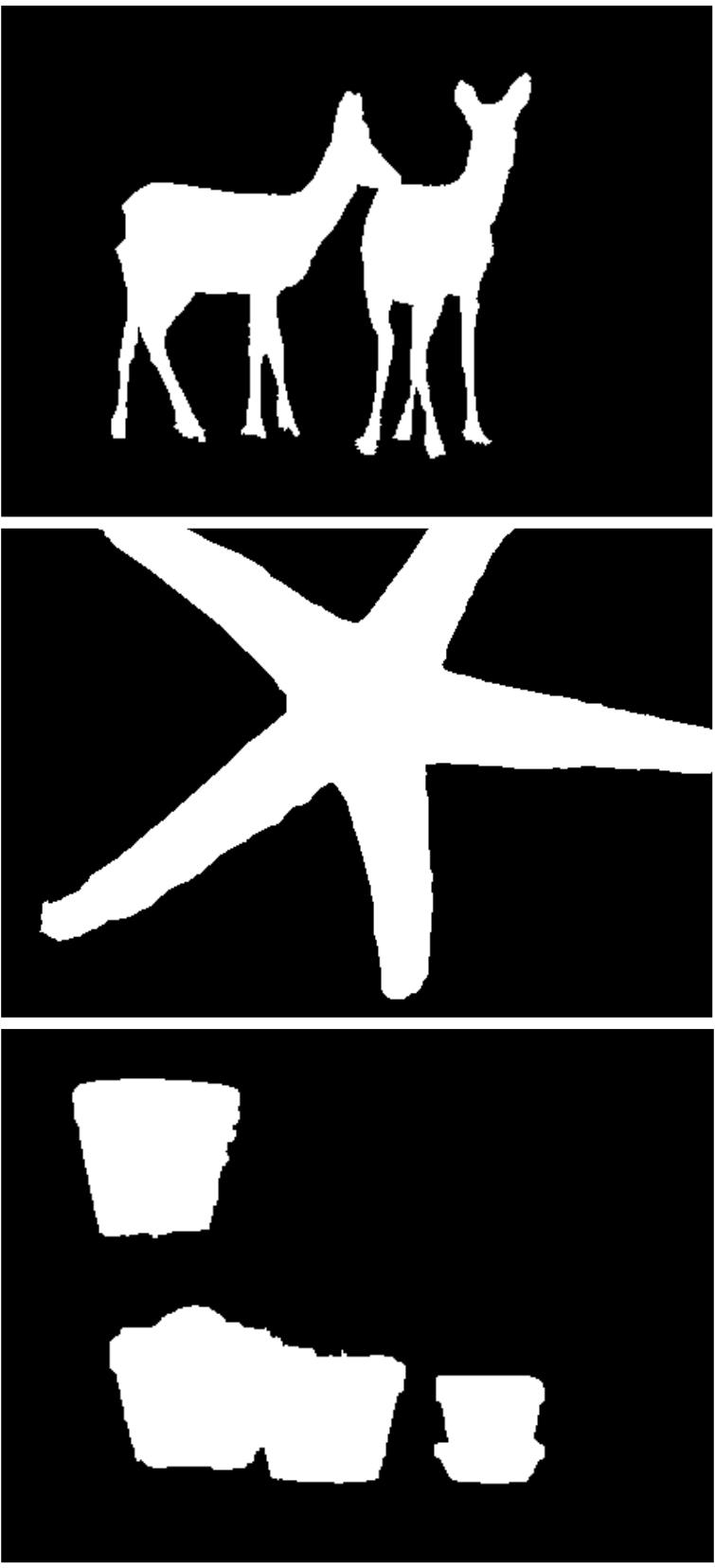}}
 	\subfigure[]{\includegraphics[width=2.1cm]{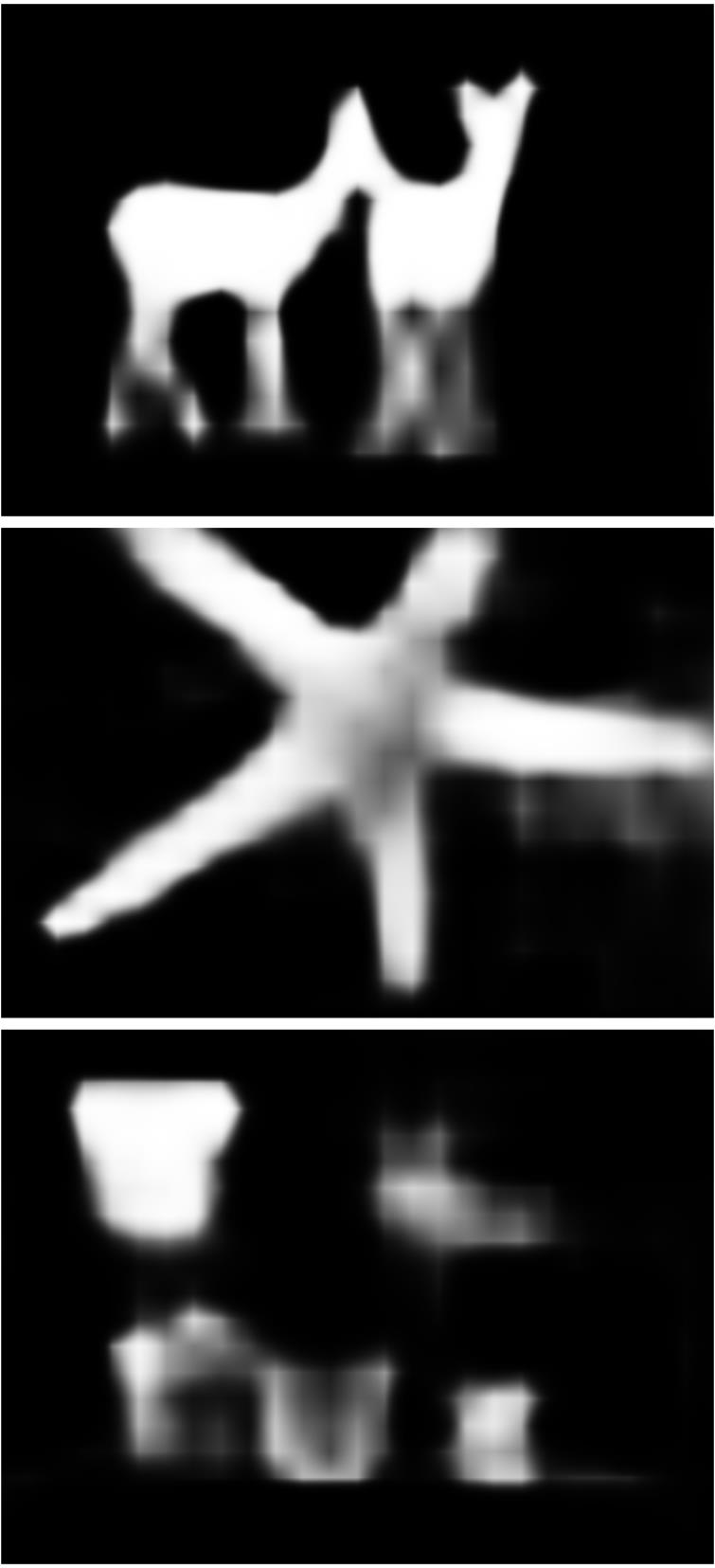}\label{f3c}}
 	\subfigure[]{\includegraphics[width=2.1cm]{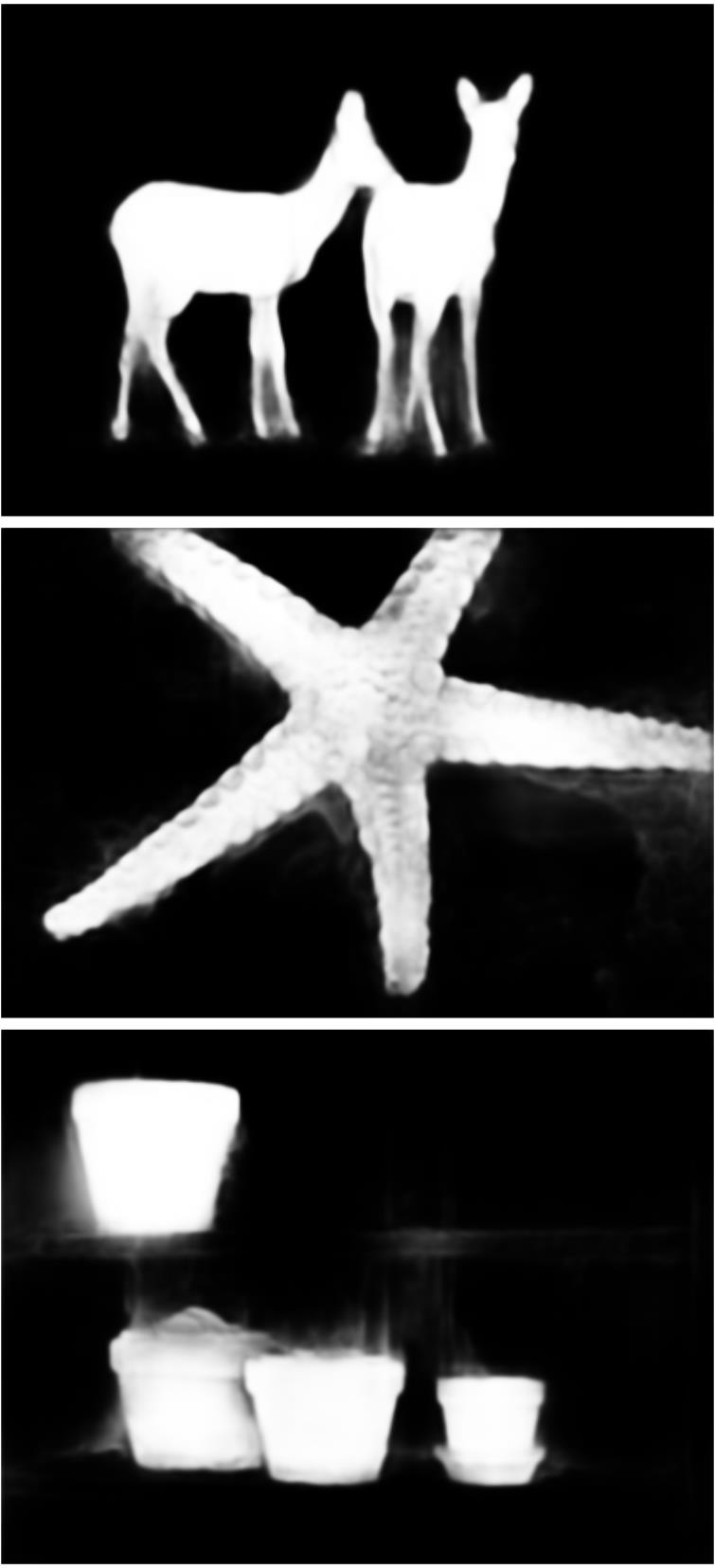}\label{f3d}}
 	\caption{Visual comparison on saliency detection results from different models for observing the effectiveness of dense connections. (a) and (b) display the input images and corresponding ground truth. (c) and (d) are the saliency detection results from the encoder model without and with dense connections.\label{fig3}}
 \end{figure}
 
As shown in the top panel of Figure \ref{network}, the non-linear transformation $G_i$ consists of two $3{\times}3$ convolutional layers. Such architecture design is based on following considerations. Using a single trainable layer to build a model with sufficiently large capacity is not easy~\cite{he2016identity}. Therefore,  we stack two $3{\times}3$ convolution layers in the context block for  stronger learning capacity. Combining the context block and dense connections, the proposed saliency encoder within SENet can exploit the supportive clues from multi-level features effectively. In particular, high-level semantic features would help localize the salient object while the low-level fine-grained features would refine the details (\eg boundaries).

Figure~\ref{fig3}   illustrates how the dense connections benefit  salient object detection  through comparing results with the one from the model without dense connections. For getting the results without dense connections in Figure~\ref{f3c}, we degenerate the architecture specified in  Eqn.~\eqref{eq3} into $x_{i-1}\leftarrow [x_{i-1}, H^{\mathrm{up}}_i(x_i)]$. Namely, only a single adjacent connection is used. Comparing results in  Figure~\ref{f3c} and Figure~\ref{f3d}, one can observe that the dense connections cannot only sharpen the object boundary (\eg, the legs of the deer) but also provide complete object regions (\eg, the plant pots). These desired benefits clearly show the advantages of dense connections for allowing direct access of classifier to low- and high-level features.

\subsection{Self-explanatory Generator}
Though we have carefully designed the encoder, it  presents little transparency for its internal mechanism on saliency detection. It also lacks justification on how the saliency prediction is made. To address this issue, given an image of interest, we propose a generator to produce an explicit explanation map that quantizes  sensitivity of the saliency encoder over each image location thoroughly. Through analyzing the generated explanation map, we can effectively reveal the important factors that influence the saliency predictions of the encoder, providing explanations on  the encoder prediction and justifying the results.

The purpose of the self-explanatory generator is to explicitly quantize how each element of the input image (or feature) $ \mathbf{y} = \{y_1,\ldots,y_M\} $ contributes to the final saliency prediction. Let $y_m$ denote the investigated feature element, \eg one patch of the input image. Here $ m $ is the feature element index.
Inspired by the work of \cite{robnik2008explaining} and \cite{zintgraf2017visualizing}, for analyzing the effect of  feature $y_m$, the generator purposively keeps it from contributing to the encoder  and only passes the other features $ \mathbf{y}_{\backslash m} $ (\eg, the other patches of the input image) to the encoder. The output is taken  as  prediction of the generator. Compared to the encoder prediction using all features $\mathbf{y}$, the prediction of the generator reflects the sensitiveness of encoder over feature $y_m$. After all the features within $\mathbf{y}$ (that denotes the whole image) are examined by the generator, the integral results would provide interpretation on how the encoder utilizes input $\mathbf{y}$ for saliency detection.

In particular, we build the generator as a probabilistic model that specifies a saliency prediction $p$ with the absence of an investigated feature $y_m$:
\begin{equation}\label{eq4}
{\rm gen} (i\lvert \mathbf{y}_{\backslash m}) \equiv p(i\lvert \mathbf{y}_{\backslash m}), 
\end{equation}
where $p(i \lvert y) $ is the predicted saliency probability with features $y$ at the location $i$, and  $\mathbf{y}_{\backslash m}$ denotes the set of all features $\mathbf{y}$ excluding $y_m$.

To leave $y_m$ out cleanly from $\mathbf{y}$ (note it is not equivalent to naively zeroing $y_m$), we need to give an unknown label to $y_m$ and retrain the saliency classifier to ignore the contribution of $y_m$ over the saliency prediction. However, the cost of re-training a DNN model is too expensive and  clearly infeasible for explaining the DNN based saliency detection models. Another approach is to simulate the absence of $y_m$ by applying the following Bayesian rules~\cite{robnik2008explaining}:
\begin{equation}\label{eq5}
p(i\lvert \mathbf{y}_{\backslash m}) = \sum_{y_m}p(y_m\lvert \mathbf{y}_{\backslash m})p(i\lvert \mathbf{y}_{\backslash m}, y_m) , 
\end{equation}
where $p(i\lvert \mathbf{y}_{\backslash m}, y_m) = p(i\lvert \mathbf{y})$  is the actual saliency prediction of the encoder at the location $i$. Because $y_m$ is assumed to be unknown, all possible values of $y_m$ need to be considered.

\begin{figure}[t]
	\begin{center}
		\includegraphics[width=1.0\linewidth]{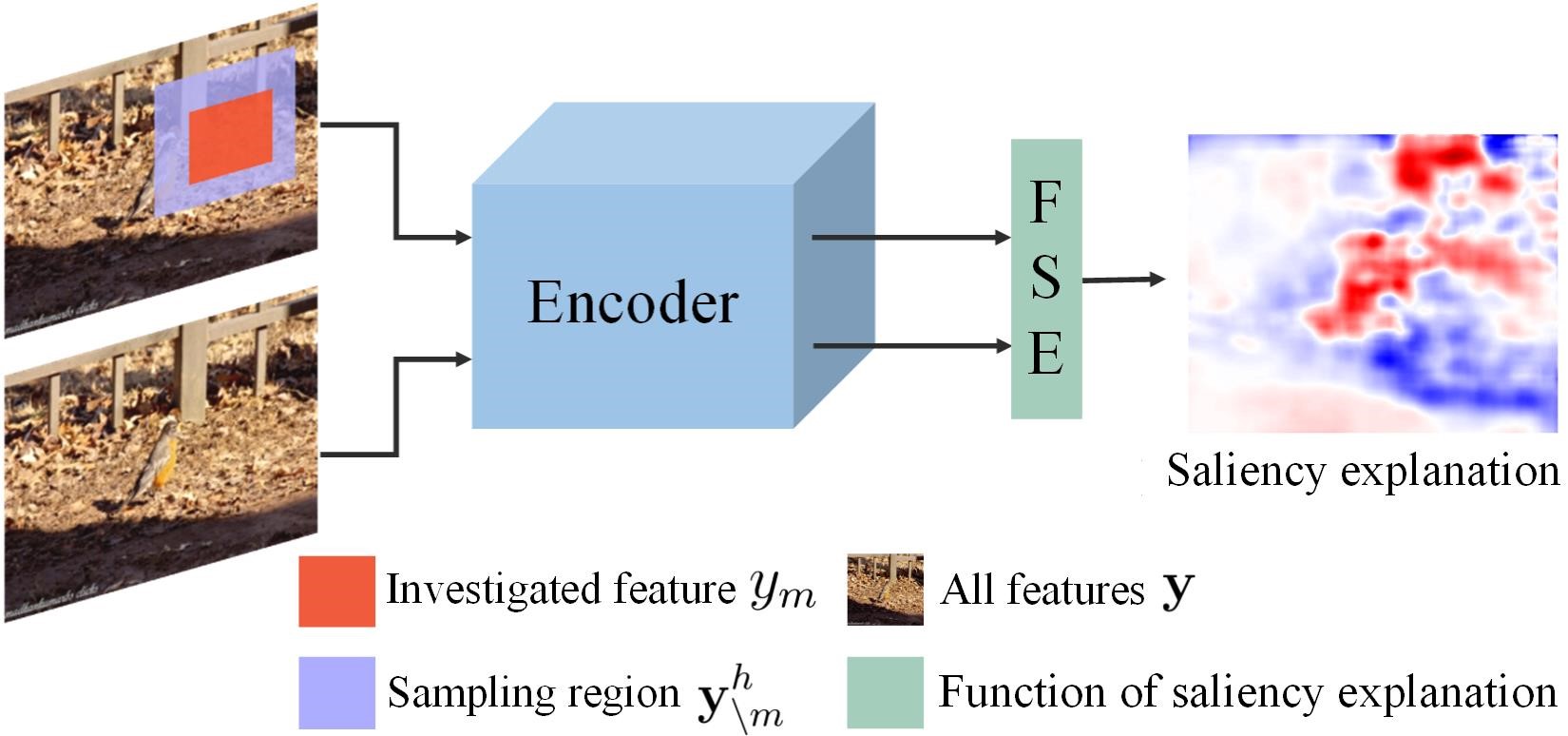}
	\end{center}
	\caption{Diagram of the proposed self-explanatory generator within SENet.  For the given image $\mathbf{y}$, the generator approximately estimates importance  $p(y_m\lvert \mathbf{y}_{\backslash m})$ for each patch $ y_m $ by sampling within the neighborhood $y_m$ (highlighted in purple in the left). Then the features $\mathbf{y}$ and $\mathbf{y}_{\backslash m}$ are passed to the encoder, giving  the saliency prediction $p(i\lvert \mathbf{y})$ and $p(i\lvert \mathbf{y}_{\backslash m})$ respectively. The final explanatory evidence is generated by the function of saliency explanation (FSE). See main text for more details. Best viewed in color.}
	\label{gen}
\end{figure}

How to estimate $p( y_m \lvert \mathbf{y}_{ \backslash m} ) $ in Eqn.~\eqref{eq5} is crucial for the generator to interpret the contribution of $ y_m $. During the image saliency detection, each image pixel can be regarded as a feature $y_m$ which however makes the estimation of $p(y_m\lvert \mathbf{y}_{\backslash m})$  impossible due to huge search space. In \cite{robnik2008explaining}, $y_m$ is directly assumed to be independent of $\mathbf{y}_{\backslash m}$, and the $p(y_m\lvert \mathbf{y}_{\backslash m})$ can be simplified as follows:
\begin{equation}\label{eq6}
p(y_m\lvert \mathbf{y}_{\backslash m}) \approx p(y_m).
\end{equation}
The above  approximation is simple but not accurate as it ignores the strong dependence among spatially  neighboring pixels within an image. As suggested by～\cite{zintgraf2017visualizing}, for a given pixel $y_m$, multiple image patches $\mathbf{y}^{h}$ around it can be sampled to approximate $\mathbf{y}$ where $h$ denotes the neighboring regions of $y_m$. Hence,  as shown in the left of  Figure \ref{gen}, we approximate $p(y_m\lvert \mathbf{y}_{\backslash m})$ by sampling patches within  $y_m$ neighborhood:
\begin{equation}\label{eq7}
p(y_m\lvert \mathbf{y}_{\backslash m}) \approx p(y_m\lvert \mathbf{y}^{h}_{\backslash m}).
\end{equation}
Once obtaining estimate of  $p(y_m\lvert \mathbf{y}_{\backslash m})$, the generator prediction  ${\rm gen} (i\lvert \mathbf{y}_{\backslash m})$ is computed  as follows:
\begin{equation}\label{eqn7}
{\rm gen} (i\lvert \mathbf{y}_{\backslash m}) \approx \sum_{y_m}p(y_m\lvert \mathbf{y}^h_{\backslash m})p(i\lvert \mathbf{y}).
\end{equation}

We can compare  outputs of the generator and the encoder to explain reaction of the encoder at feature $y_m$. A comparison function, called saliency explanation, is defined to measure the sensitiveness of the proposed saliency encoder on investigated feature $y_m$: 
\begin{align}\label{eq8}
{\rm FSE}(i \lvert y_m) &= {\rm diff} ({\rm enc}(i \lvert \mathbf{y}), {\rm gen} (i\lvert \mathbf{y}_{\backslash m})) \nonumber \\ 
&= \log \frac{{\rm enc}(i \lvert \mathbf{y})}{1-{\rm enc}(i \lvert \mathbf{y})} - \log \frac{{\rm gen}(i\lvert \mathbf{y}_{\backslash m})}{1-{\rm gen}(i\lvert \mathbf{y}_{\backslash m})},
\end{align}
where ${\rm enc}(i \lvert \mathbf{y})$ means the encoder prediction at location $i$ over feature $\mathbf{y}$.

The ${\rm FSE}(i \lvert y_m)$ in Eqn.~\eqref{eq8} reflects how the feature $y_m$ influences the saliency prediction of the encoder on the $ i $-th location. ${\rm FSE}(i \lvert y_m){>}0$ means removing $y_m$ causes the decreasing of the saliency score and $y_m$ thus has a positive influence on the saliency prediction of the encoder. ${\rm FSE}(i \lvert y_m){<}0$ means removing $y_m$ brings a increasing saliency score. That is, the presence of $y_m$ may suppress  the saliency prediction of the encoder. Using Eqn.~\eqref{eq8}, we get the explanation map ${\rm FSE}(y_m)$ of feature $y_m$ at each position $i$. After going through all features $\mathbf{y}$, we can get explanation maps for all the used  features. To integrate them together, we sum and average all the evidence maps into the explanatory ${\rm FSE}(\mathbf{y})$ for $\mathbf{y}$. 

In this paper, we focus on  explaining dependence of  the saliency model on the input image $I$. Then ${\rm FSE}(I)$ is adopted for explanation. Through visualizing ${\rm FSE}(I)$, we  get supportive saliency evidence map for the proposed saliency encoder. Note the self-explanatory generator can be applied for explaining any DNN based saliency detection model besides the proposed saliency encoder model in this paper. Hence, ${\rm FSE}(I)$ can be regarded as a metric to compare the capability of different models. More powerful the saliency detection model is, more reasonable the ${\rm FSE}(I)$ should be. We believe such an explanation model is  of independent interest to the computer vision community to better understand predictions of ``black-box'' DNN models.

\section{Experiments}\label{sec4}
\subsection{Datasets \label{4.1}}
In this section, we evaluate the performance of the proposed SENet model for salient object detection on five widely used benchmark datasets: MSRA-B~\cite{liu2011learning}, ECSSD \cite{shi2016hierarchical}, HKU-IS~\cite{li2015visual}, PASCAL-S \cite{li2014secrets} and SOD \cite{martin2001database}. The MSRA-B dataset~\cite{liu2011learning} contains 5,000 images from hundreds of different categories and most images  only have one object instance. ECSSD \cite{shi2016hierarchical} contains 1,000 semantically meaningful but structurally complex images. HKU-IS \cite{li2015visual} is one of the latest datasets and contains more than 4,000 challenging images, including low contrast or multiple salient objects. PASCAL-S \cite{li2014secrets} is built on the validation set of PASCAL VOC 2012 segmentation challenge which contains 850 natural images. Due to complex object context and cluttered background, PASCAL-S \cite{li2014secrets} is one of the most challenging salient object detection datasets. SOD \cite{martin2001database} contains 300 images, most of which include multiple salient instances. All the datasets provide pixel-level human annotations on salient object.

\subsection{Evaluation Metrics}
Three widely used and standard metrics are adopted to evaluate our model, \ie,  precision-recall (PR) curves, F-measure, and the mean absolute error (MAE). The PR curves depict the mean precision and recall of all testing images against the ground truth. The mean precision and recall of an image are computed by polarizing the saliency map with a  threshold $T_f$ varying within the range of $[0\mathbin{:}1\mathbin{:}255]$ and comparing results  with the ground truth. Similar to~\cite{achanta2009frequency}, an adaptive threshold $T_a$ is employed to compute the weighted F-measure. The adaptive threshold $T_a$ is defined as proportional to the mean of the saliency map $S$ as follows:
\begin{equation}\label{eq21}
T_a = \frac{t}{W\times H} \sum_{i=1}^W\sum_{j=1}^HS(i,j),
\end{equation}
where $t$ is typically set to be 1.5 \cite{ss22}. Here $H$ and $W$ are the height and width of the image respectively.

The binary map generated by using the threshold of  $T_a$ is employed to compute the precision, recall and weighted F-measure:
\begin{equation}\label{eqfm}
F_\omega = \frac{(1+\omega^2)\mathrm{Precision}\times \mathrm{Recall}}{\omega^2\times \mathrm{Precision} + \mathrm{Recall}},
\end{equation}
where $\mathrm{Precision}$ and $\mathrm{Recall}$ denote the proportion of detected true positive saliency pixels as compared to the number of detected saliency pixels and the number of saliency pixels in the ground truth, respectively. The parameter $\omega^2$ is set to 0.3 for emphasizing the precision~\cite{achanta2009frequency}.

Finally, MAE is defined as the mean absolute difference between the saliency map $S$ and the ground truth $G$:
\begin{equation}\label{eqmae}
{\rm MAE} = \frac{1}{H\times W}\sum_{i=1}^{H}\sum_{j=1}^{W}\left| S(i,j)-G(i,j)\right|.
\end{equation}
MAE considers the true negative rates while the $F_\omega$ focuses on the successful detected regions \cite{cheng2013efficient}.
\begin{table}[t]
	\centering
	\caption{Layer configuration of context block.}\label{tab2}
	\begin{tabular}{clccc}
		\toprule
		\multirow{2}{*}{Stage} & \multicolumn{2}{c}{ResNet-101} & \multicolumn{2}{c}{VGG-16} \\
		\cmidrule(lr){2-3} \cmidrule(lr){4-5}
		& Layer & Kernel & Layer & Kernel \\
		\midrule
		1 & - & - & {\ttfamily conv1$\_$2} & $\left[\begin{array}{l}\textsc{3$\times$3, 32} \\ \textsc{3$\times$3,  32}  \end{array} \right] $ \\
		\midrule
		2 & {\ttfamily res2c} & $\left[\begin{array}{l}\textsc{3$\times$3, 256} \\ \textsc{3$\times$3,  256}  \end{array} \right] $ & {\ttfamily conv2$\_$2} & $\left[\begin{array}{l}\textsc{3$\times$3, 64} \\ \textsc{3$\times$3, 64}  \end{array} \right] $    \\ 
		\midrule
		3 & {\ttfamily res3b3} & $\left[\begin{array}{l}\textsc{3$\times$3, 256} \\ \textsc{3$\times$3,  256}  \end{array} \right] $ & {\ttfamily conv3$\_$3} & $\left[\begin{array}{l}\textsc{3$\times$3, 128} \\ \textsc{3$\times$3, 128}  \end{array} \right] $  \\
		\midrule
		4 & {\ttfamily res4b22} & $\left[\begin{array}{l}\textsc{3$\times$3, 256} \\ \textsc{3$\times$3,  256}  \end{array} \right] $ & {\ttfamily conv4$\_$3} & $\left[\begin{array}{l}\textsc{3$\times$3, 256} \\ \textsc{3$\times$3,  256}  \end{array} \right] $ \\ 
		\midrule
		5 & {\ttfamily res5c} & $\left[\begin{array}{l}\textsc{3$\times$3, 512} \\ \textsc{3$\times$3,  512}  \end{array} \right] $ & {\ttfamily conv5$\_$3} & $\left[\begin{array}{l}\textsc{3$\times$3, 256} \\ \textsc{3$\times$3,  256}  \end{array} \right] $  \\
		\bottomrule
	\end{tabular}
\end{table}

\begin{table*}[t]
	\renewcommand{\arraystretch}{1.3}
	\centering
	\caption{Quantitative comparison with state-of-the-arts. The up-arrow $\uparrow$ means  larger is better while the down-arrow $\downarrow$ means smaller is better. Symbol $\dagger$ indicates that the network is pre-trained on ResNet-101 \cite{he2016deep} while No $\dagger$ indicates that the pre-trained model is VGG-16 \cite{simonyan2014very}. For the VGG-16 based models, the top three results are highlighted in \RB{red}, \GB{green} and \BB{blue}. \label{tab1}}
	\begin{threeparttable}
		\begin{tabular}{l|c|c|c|c|c|c|c|c|c|c}
			\hline
			\multirow{2}{*}{\diagbox{Methods}{Datasets}} & \multicolumn{2}{c|}{MSRA-B～\cite{liu2011learning}} & \multicolumn{2}{c|}{ECSSD～\cite{shi2016hierarchical}} & \multicolumn{2}{c|}{HKU-IS～\cite{li2015visual}} & \multicolumn{2}{c|}{PASCAL-S～\cite{li2014secrets}} & \multicolumn{2}{c}{SOD～\cite{martin2001database}} \\ \cline{2-11}
			& $F_\omega \uparrow$ & MAE$\ \downarrow$ & $F_\omega \uparrow$ & MAE$\ \downarrow$ & $F_\omega \uparrow$ & MAE$\ \downarrow$ & $F_\omega \uparrow$ & MAE$\ \downarrow$ & $F_\omega \uparrow$ & MAE$\ \downarrow$ \\ \hline
			MDF～\cite{li2015visual} & 0.885 & 0.104 & 0.833 & 0.108 & 0.860 & 0.129 & 0.764 & 0.145 & 0.785 & 0.155 \\ 
			MC~\cite{zhao2015saliency} & 0.872 & 0.062 & 0.822 & 0.107 & 0.781 & 0.098 & 0.721 & 0.147 & 0.708 & 0.184 \\
			DS~\cite{li2016deepsaliency} & - & - & 0.810 & 0.160 & - & - & 0.818 & 0.170 & 0.781 & 0.150 \\
			DCL~\cite{li2016deep} & 0.916 & 0.047 & 0.898 & 0.071 & 0.907 & \BB{0.048} & 0.822 & 0.108 & 0.832 & \BB{0.126} \\
			ELD~\cite{lee2016deep} & 0.914 & \BB{0.042} & 0.865 & 0.098 & 0.844 & 0.071 & 0.767 & 0.121 & 0.760 & 0.154 \\
			DHS~\cite{liu2016dhsnet} & - & - & 0.905 & \BB{0.061} & 0.892 & 0.052 & 0.820 & \BB{0.091} & 0.823 & 0.127 \\
			RFCN～\cite{wang2016saliency} & \GB{0.926} & 0.062 & 0.898 & 0.097 & 0.895 & 0.079 & 0.827 & 0.118 & 0.805 & 0.161 \\
			CRPSD～\cite{tang2016saliency} & - & - & \RB{0.919} & 0.098 & \BB{0.911} & 0.091 & \BB{0.830} & 0.114 & - & - \\
			DLS～\cite{hu2017deep} & - & - & 0.852 & 0.086 & 0.835 & 0.091 & 0.752 & 0.120 & - & - \\
			NLDF~\cite{luo2017non} & 0.911 & 0.069 & 0.905 & 0.093 & 0.902 & 0.075 & 0.824 & 0.120 & \BB{0.840} & 0.168  \\
			DSS～\cite{hou2016deeply} & \GB{0.926} & \GB{0.029} & \GB{0.915} & \GB{0.055} & \GB{0.912} & \GB{0.040} & \GB{0.831} & \GB{0.083} & \GB{0.842} & \GB{0.121} \\
			SENet & \RB{0.931}  & \RB{0.026}  & \GB{0.915} & \RB{0.052} & \RB{0.924} & \RB{0.032} & \RB{0.840} & \RB{0.071} & \RB{0.851} & \RB{0.119}\\
			\hline
			DeepLab-v2$^\dagger$~\cite{chen2016deeplab} & 0.951 & 0.021 & 0.935 & 0.040 & 0.928 & 0.038 & 0.857 & 0.074 & 0.869 & 0.101 \\
			SENet$^\dagger$ & \textbf{0.957} & \textbf{0.020} & \textbf{0.945} & \textbf{0.036} & \textbf{0.935} & \textbf{0.032} & \textbf{0.865} & \textbf{0.066} & \textbf{0.877} & \textbf{0.095} \\
			
			\hline
		\end{tabular}
	\end{threeparttable}
\end{table*}

\subsection{Implementation Details}
Our proposed SENet is implemented by the publicly available Caffe library \cite{jia2014caffe}. We choose ResNet-101 \cite{he2016deep} and VGG-16 \cite{simonyan2014very} pre-trained on ImageNet \cite{deng2009imagenet} as our basic models. For ResNet-101, the entire model is trained end-to-end. The feature maps from {\ttfamily res2c}, {\ttfamily res3b3}, {\ttfamily res4b22} and {\ttfamily res5c} are selected and passed to the context block. Details of layer configuration for the context block are provided in Table \ref{tab2}. Due to the capability limitation, the end-to-end training is infeasible for VGG-16 based model. We split the training of VGG-16 based saliency model in two steps. Firstly, we train the VGG-16 like FCNs \cite{long2015fully} with abandoning the layers of {\ttfamily pool5}, {\ttfamily fc6} and {\ttfamily fc7}. Then the context blocks and the dense short- and long- connections are added into the per-trained model. The feature maps from {\ttfamily conv1$\_$2}, {\ttfamily conv2$\_$2}, {\ttfamily conv3$\_$3}, {\ttfamily conv4$\_$3} and {\ttfamily conv5$\_$3} are fed forward to the context block. The learning rate of pre-trained layers is set to $1/1000$ of the newly added layers. 

To ensure a fair comparison with existing approaches, we adopt a  training set of 5000 images following  the same setting in~\cite{li2015visual}: 2500 images from the MSRA-B dataset and 2500 images from the HKU-IS. All the model hyperparameters are tuned over a separate validation set containing  1000 images: 500 images from the MSRA-B dataset and the other 500 images from the HKU-IS. We evaluate the trained model over the datasets mentioned in Sec. \ref{4.1}. Momentum and weight decay parameters  are fixed as 0.9 and 0.0005 respectively. We use following  data augmentation: random mirror and random resize between 0.5 and 1.5. Augmenting the data reduces the over-fitting risk for  the network  and improves the performance. For the ResNet101-based model, we train the model with a mini-batch size of 2 for 25 epochs, and set the learning rate as $1 {\times}  10^{-8}$. For the VGG16-based model, we first train the model with a mini-batch size of 15 for 10 epochs, and set the learning rate to $5 {\times}  10^{-9}$. Then we fine-tune the pre-trained model for another 20 epochs. 

For saliency explanation, we employ a crop size  of 224 because small crop size can reduce sampling complexity  and speed up the computation in Eqn.~\eqref{eqn7}. For a testing image $I$, we compute ${\rm FSE}(I)$ in Eqn. \eqref{eq8} with a 16$\times$16 image patch each time. For a testing patch, we sample its neighbors for 5 times within a 24$\times$24 surrounding region and slide the sampling window over the whole image with a stride of 2.

\subsection{Comparison with State-of-the-arts}\label{sec_comp}
We compare the proposed SENet with 11 existing DNN based salient object detection methods, including MDF \cite{li2015visual}, MC \cite{zhao2015saliency}, DS \cite{li2016deepsaliency}, DCL~\cite{li2016deep}, ELD \cite{lee2016deep}, DHS \cite{liu2016dhsnet}, RFCN \cite{wang2016saliency}, CRPSD~\cite{tang2016saliency}, DLS～\cite{hu2017deep}, NLDF~\cite{luo2017non} and DSS~\cite{hou2016deeply}. Besides, we train a ResNet-101 based DeepLab-v2 \cite{chen2016deeplab} model for salient object detection to compare with our ResNet-101 based SENet. We replace the softmax layer of DeepLab-v2 \cite{chen2016deeplab} with a sigmoid cross entropy layer for saliency prediction. Besides, fully connected conditional random  field (CRF)~\cite{koltun2011efficient} is employed for post-processing the results.
\begin{figure}[t]
	\centering
	\subfigure[ECSSD]{\includegraphics[width=4.35cm]{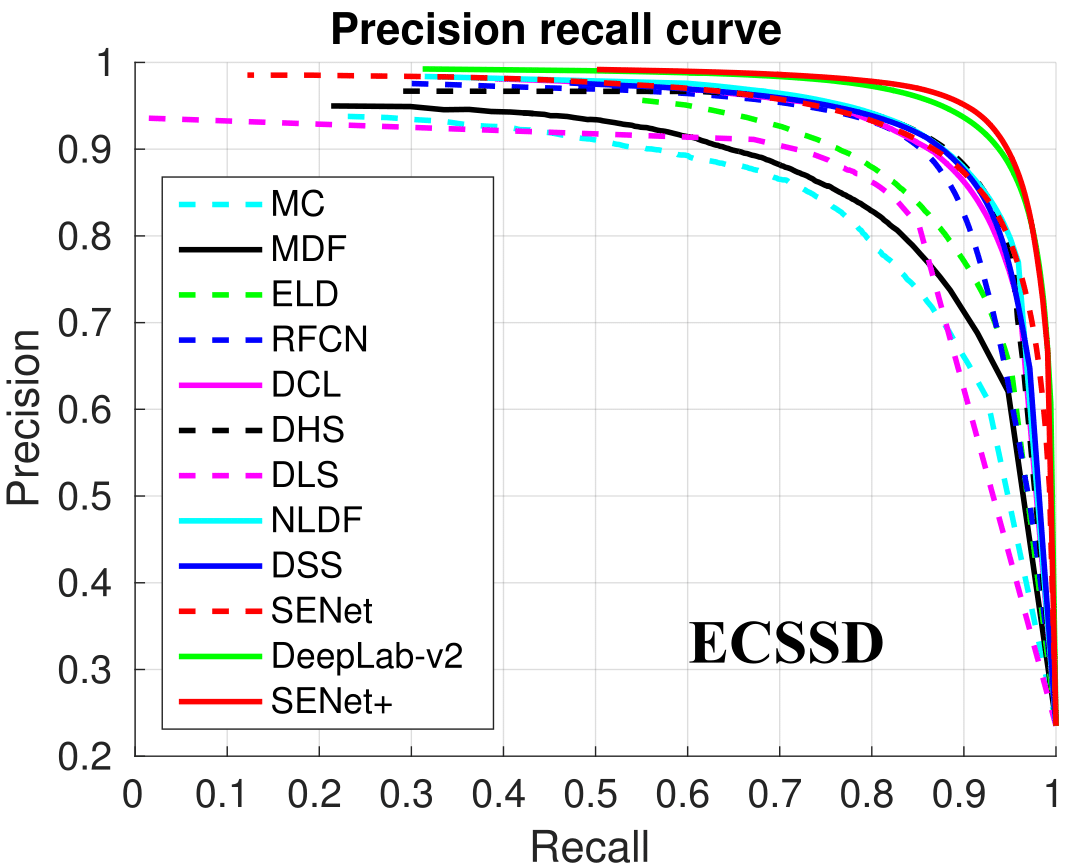}}
	\subfigure[HKU-IS]{\includegraphics[width=4.35cm]{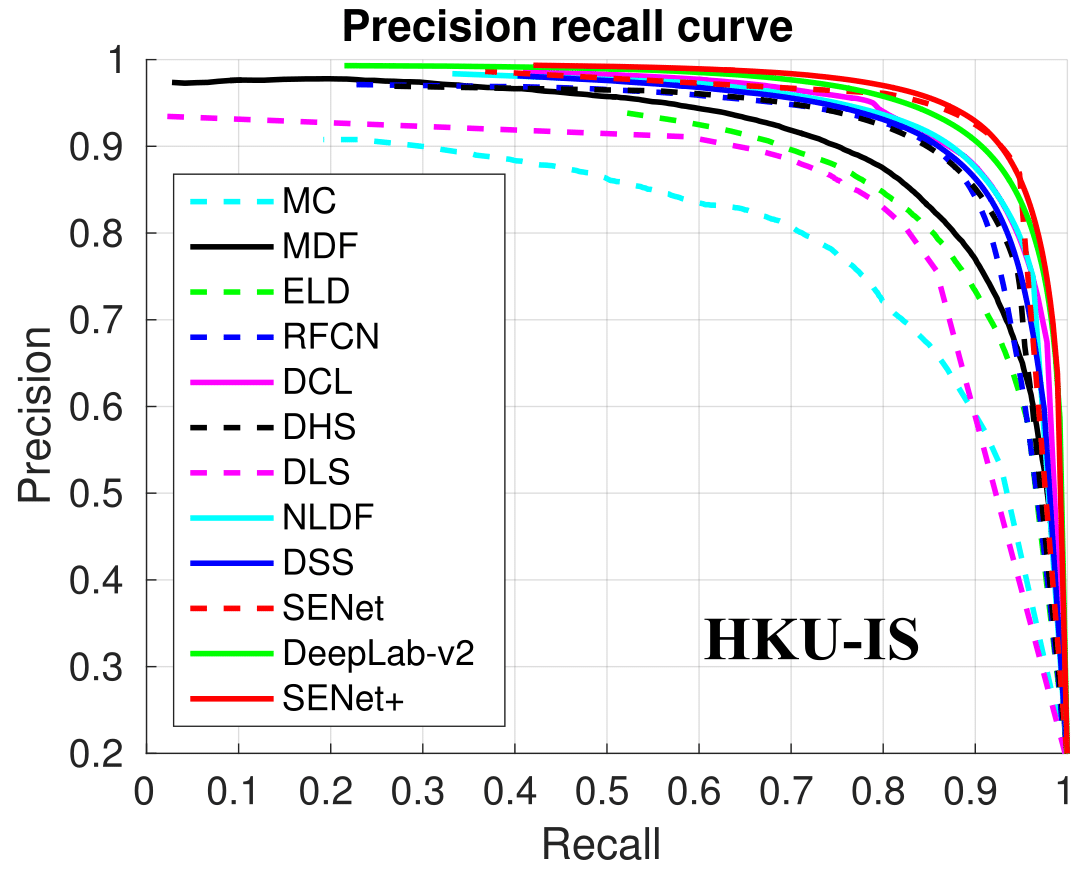}} \\
	\subfigure[PASCAL-S]{\includegraphics[width=4.35cm]{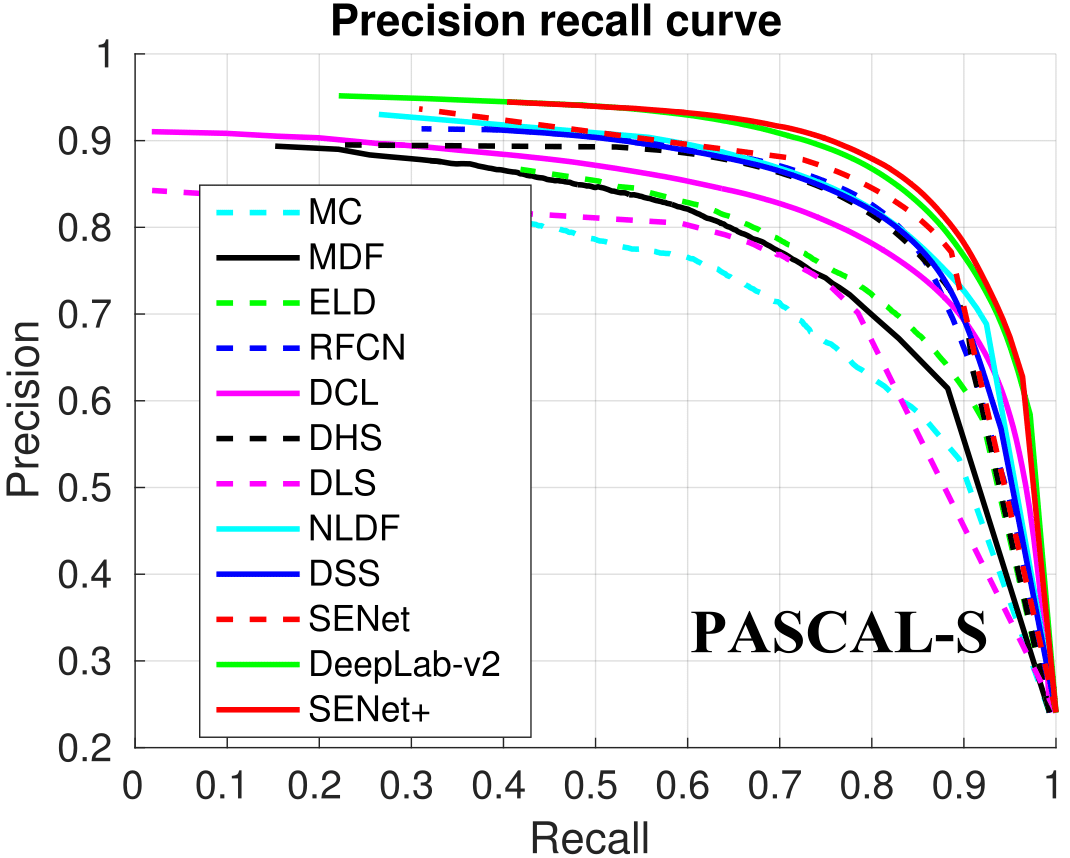}}
	\subfigure[SOD]{\includegraphics[width=4.35cm]{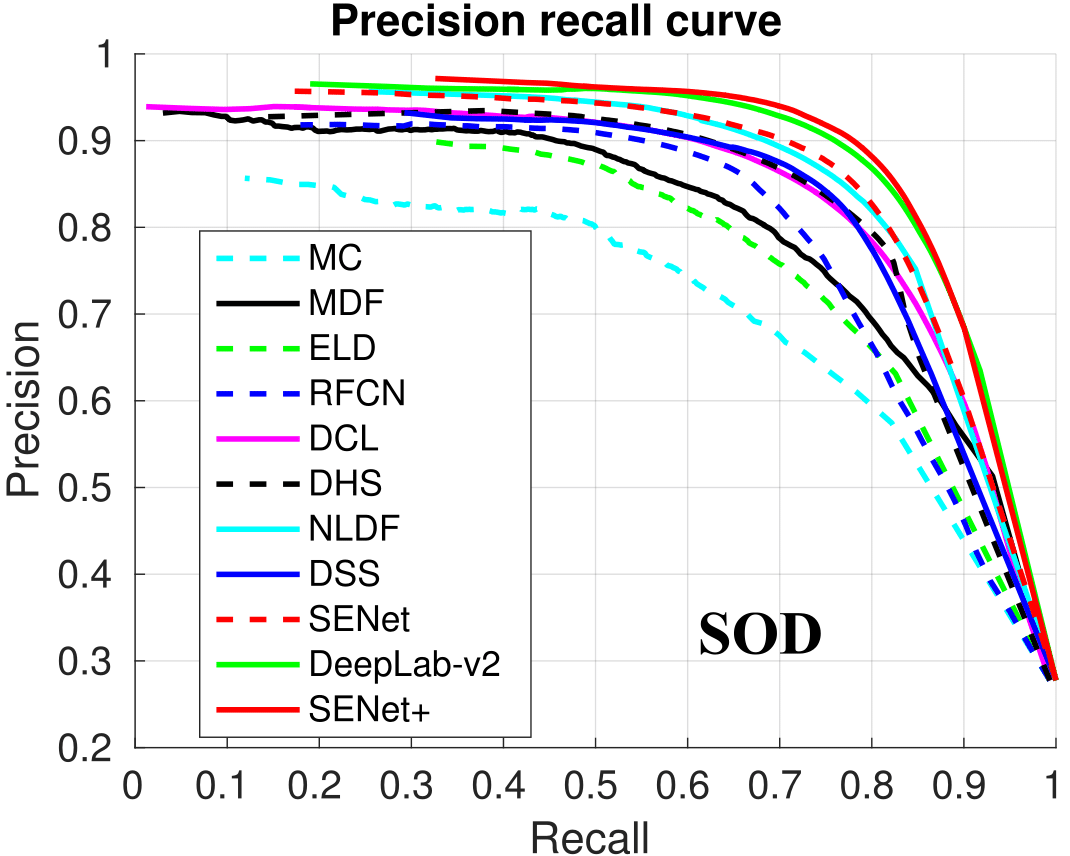}}
	\caption{Comparison of precision-recall curves of 11 DNN based salient object detection approaches on 4 datasets. Best viewed in color. \label{pr}}
\end{figure}

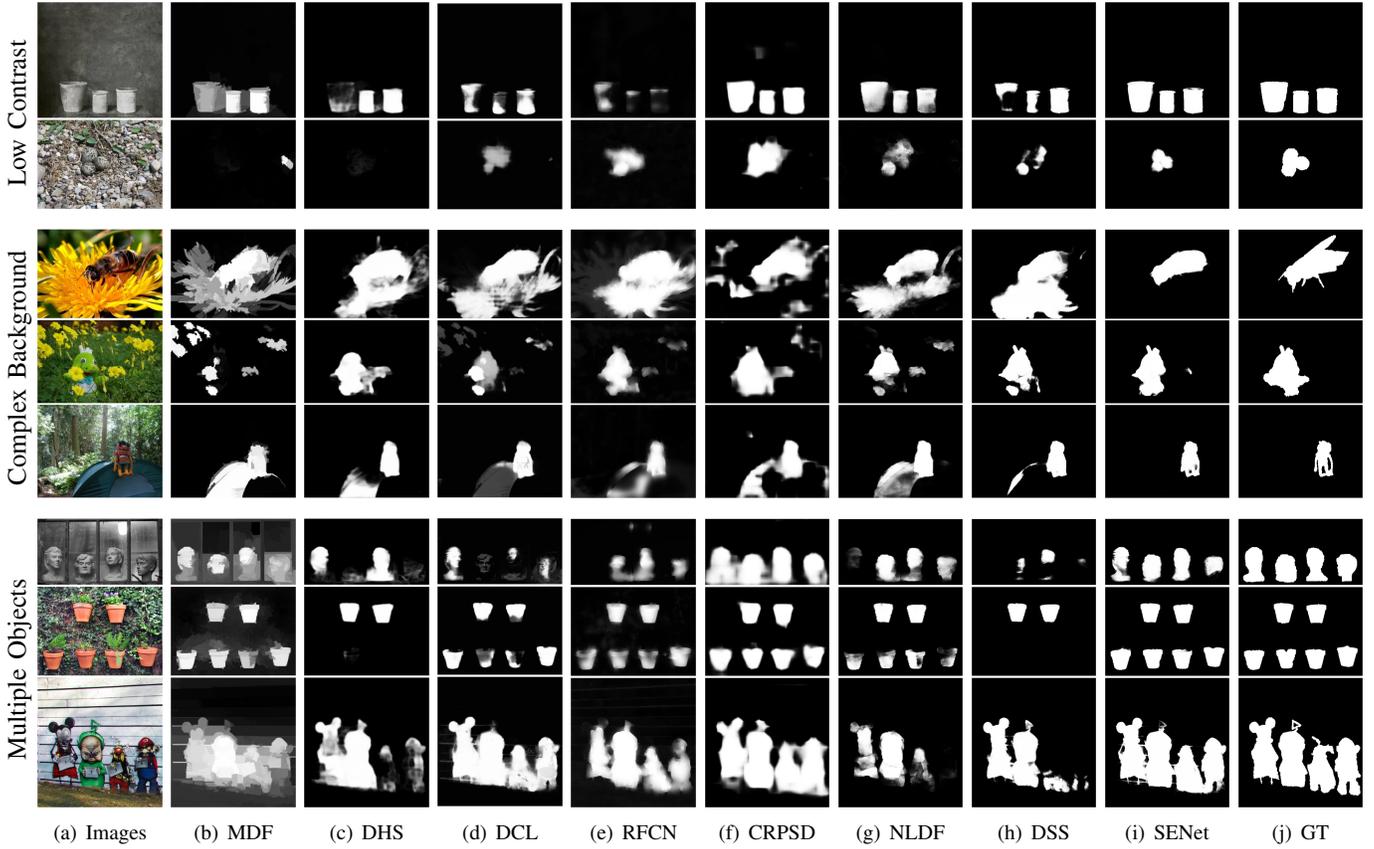
\begin{figure*}[t]
	\centering
	\subfigure[Images]{\includegraphics[width=1.65cm]{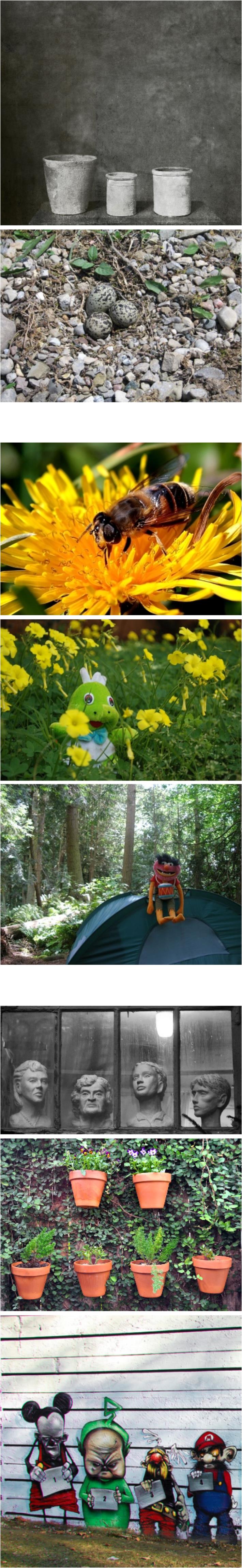}}
	\subfigure[MDF]{\includegraphics[width=1.65cm]{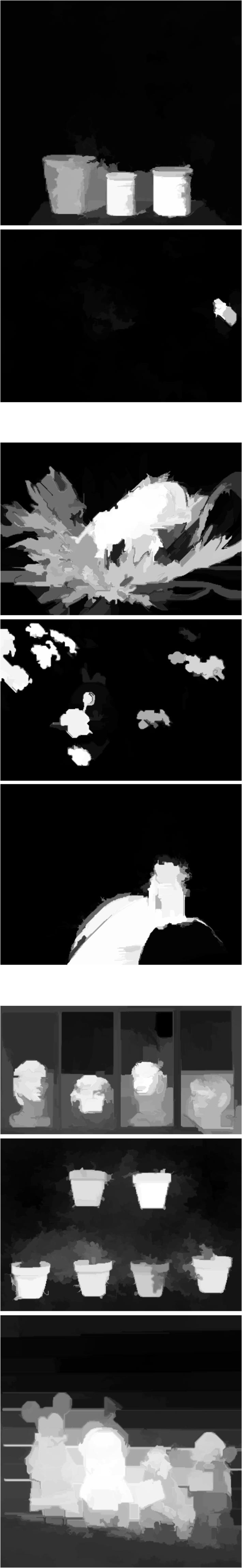}}
	\subfigure[DHS]{\includegraphics[width=1.65cm]{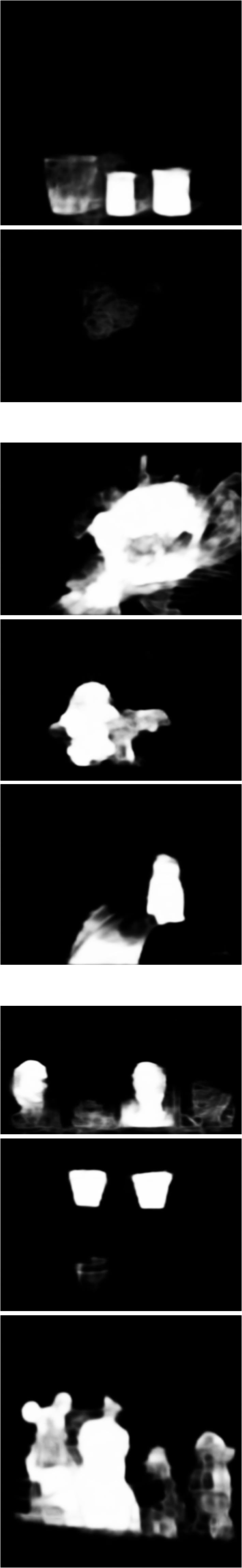}}
	\subfigure[DCL]{\includegraphics[width=1.65cm]{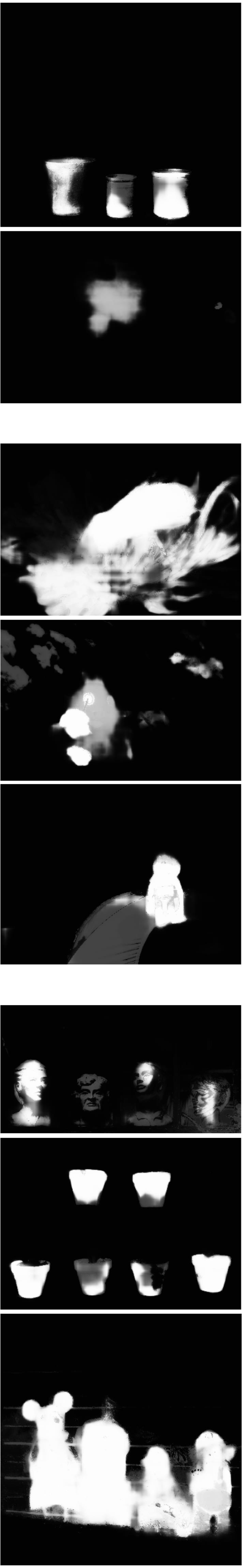}}
	\subfigure[RFCN]{\includegraphics[width=1.65cm]{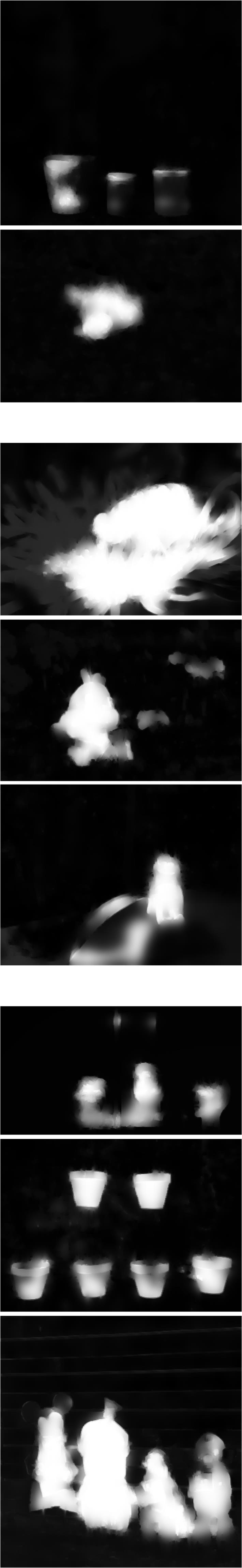}}
	\subfigure[CRPSD]{\includegraphics[width=1.65cm]{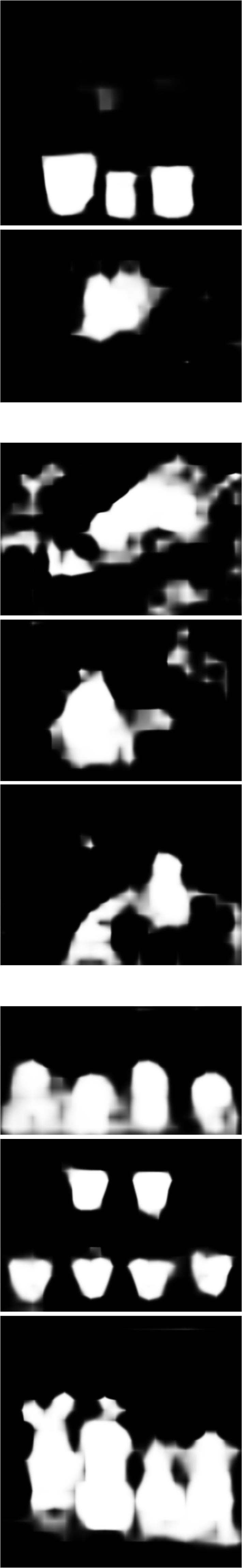}}
	\subfigure[NLDF]{\includegraphics[width=1.65cm]{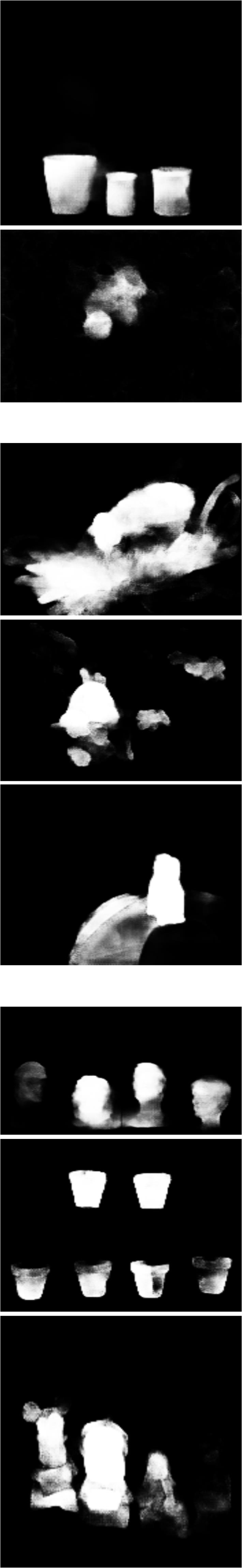}}
	\subfigure[DSS]{\includegraphics[width=1.65cm]{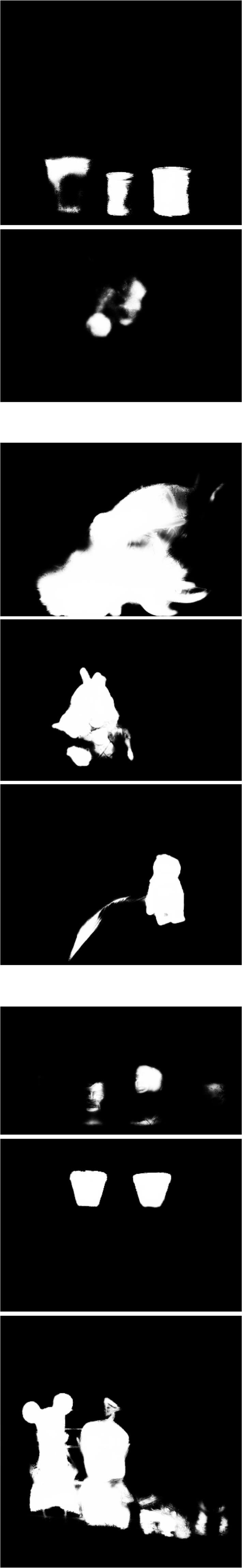}}
	\subfigure[SENet]{\includegraphics[width=1.65cm]{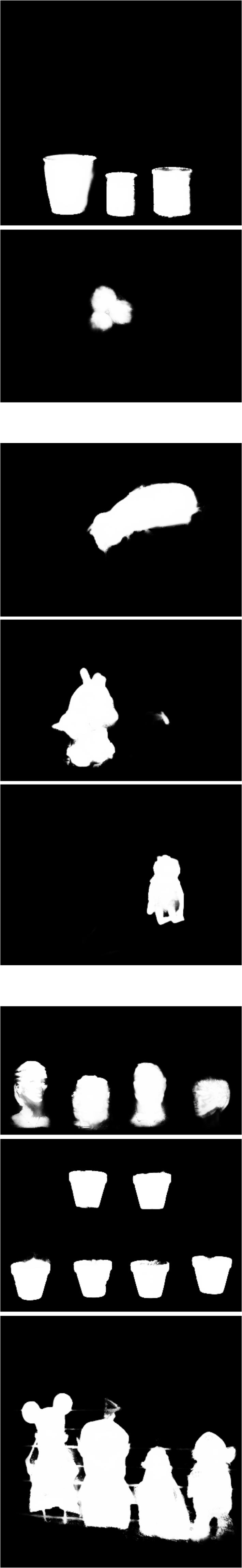}}
	\subfigure[GT]{\includegraphics[width=1.65cm]{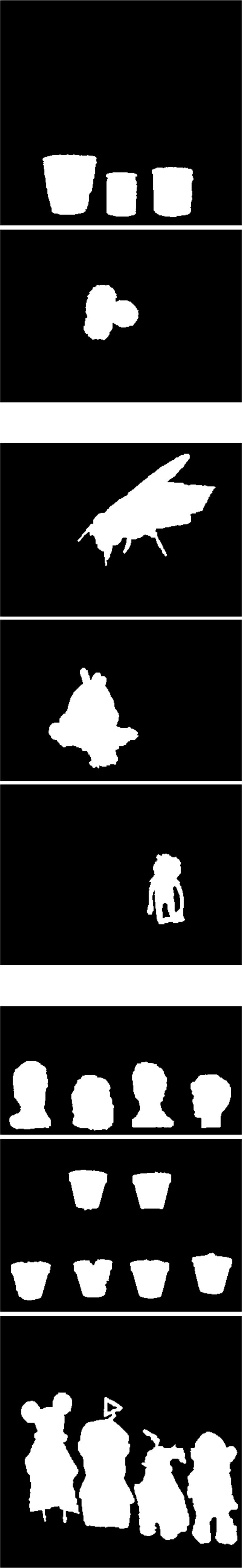}}
	\begin{picture}(0,0)
	\put(-516,235){\rotatebox{90}{Low Contrast}}
	\put(-516,122){\rotatebox{90}{Complex Background}}
	\put(-516,18){\rotatebox{90}{Multiple Objects}}
	\end{picture}
	\caption{Visual comparisons of salient object detection results. We present three typical challenging in salient object detection, including low contrast, complex background and multiple salient objects. The compared methods include: MDF~\cite{li2015visual}, DHS~\cite{liu2016dhsnet}, DCL~\cite{li2016deep}, RFCN~\cite{wang2016saliency}, CRPSD~\cite{tang2016saliency}, NLDF~\cite{luo2017non} and DSS~\cite{hou2016deeply}. The ground truth (GT) is shown in the last column. \label{viscom}}
\end{figure*}

\subsubsection{Quantitative comparison}
In Table \ref{tab1}, we quantitatively compare state-of-the-arts in term of maximum MAE and F-measure. For VGG-16～\cite{simonyan2014very} based models, the proposed SENet outperforms other compared approaches in terms of MAE, improving existing best results by 10.3$\%$, 5.4$\%$, 20.0$\%$, 14.5$\%$ and 1.7$\%$ respectively on MSRA-B, ECSSD, HKU-IS, PASCAL-S and SOD. The much smaller  MAE of SENet demonstrates that SENet can  better detect the non-salient regions and offer better true negative rates. Besides, the proposed SENet improves the F-measure of existing best-performing approaches by 0.5$\%$, 1.3$\%$, 1.1$\%$ and 1.1$\%$ respectively on MSRA-B, HKU-IS, PASCAL-S and SOD. CRPSD~\cite{tang2016saliency} achieved the best F-measure of ECSSD which outperforms SENet by 0.4$\%$. Based on the pre-trained model ResNet-101～\cite{he2016deep}, the proposed method outperforms DeepLab-v2～\cite{chen2016deeplab} by an average margin of 9.5$\%$ and 0.8$\%$  for  MAE and F-measure respectively. It is because our model  densely connects low- and high-level context which makes the classifier have global and local connections with the computed features. Though the dilated convolution~\cite{chen2016deeplab} is proposed to maintain high-resolution feature maps, the connection between classifier and feature maps is diluted by the  operation of dilation~\cite{peng2017large}. The local connected classifier loses its global perspective which could  hurt performance for detecting salient object out of the receptive field.

We show the results of PR curves in  Figure~\ref{pr}. The PR curves depict the  rates of true positive. Usually,  higher precision and slower attenuation of the curve indicate  better capability of the salient object detection model. The ResNet-101 based SENet$^\dagger$ outperforms other compared methods significantly on all the benchmark datasets. For the VGG-16 based models, SENet preserves the precision at the highest level with  increased recall. For the  HKU-IS and PASCAL-S datasets,  at high level of recall (recall$>$0.9), DCL~\cite{li2016deep} and NLDF~\cite{luo2017non} can provide better precision. Though DSS~\cite{hou2016deeply} achieves the second best performance in terms of the MAE and F-measure, NLDF~\cite{luo2017non} goes beyond DSS~\cite{hou2016deeply} under the evaluation of PR curves. 

Overall, the proposed SENet provides new state-of-the-art for DNN based salient object detection in terms of MAE, F-measure and PR curves consistently. For various benchmark datasets, such as relative simple MSRA-B or challenging PASCAL-S and SOD, the performance of the proposed SENet is impressive.

\subsubsection{Visual comparison}\label{sec_viscomp}
In Figure~\ref{viscom}, we present three typical challenging cases in salient object detection, including: low contrast, complex background and multiple salient objects. The low contrast case presents a scenario where  the salient object has similar color or texture to the background. As shown in the top two rows of Figure \ref{viscom}, the target object has similar visual features as the background which makes it hard to be segmented accurately. For this challenging case, most compared methods except CRPSD~\cite{tang2016saliency} fail to  detect the integrated salient objects. The proposed SENet detects the salient object entirely and provides sharp boundaries along the salient object. The second challenging case we consider  is from  complex background, as shown  in the middle of Figure~\ref{viscom}. The background presents complex structure which easily leads to false positive detection. For example, the \textit{ flower} in the third row of Figure \ref{viscom} has bright color which increases error for all the compared methods. For this example, the low-level features cannot work well. However, our proposed SENet can address this challenging case well. Its superior performance benefits from its internal dense connections of multi-level features and exploiting the high-level feature\textemdash the context information of the \textit{ bee}\textemdash to detect the salient object. The last challenging scenario presents multiple salient objects. Some models, such as DHS~\cite{liu2011learning}, DCL~\cite{li2016deep}, NLDF~\cite{luo2017non} and DSS~\cite{hou2016deeply} severely miss parts of the salient object while  others (e.g., MDF~\cite{li2015visual}) incorrectly include the background regions into the detection results. By contrast, the proposed SENet detects each object instance accurately.

\begin{figure}[t]
	\centering
	\subfigure[Trimap examples]{\includegraphics[width=4.2cm, height=3.2cm]{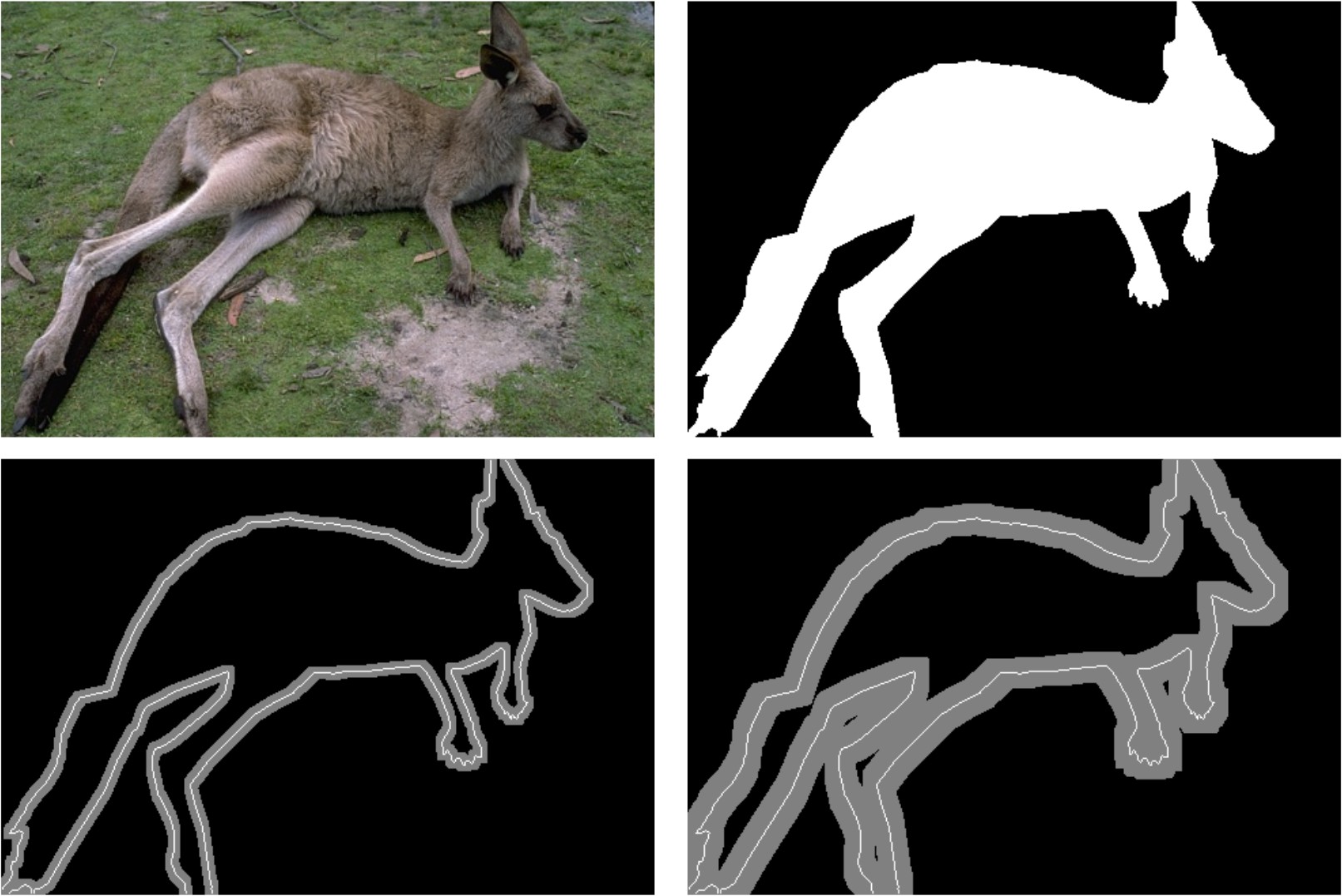}\label{bnd_a}}
	\subfigure[HKU-IS]{\includegraphics[width=4.35cm]{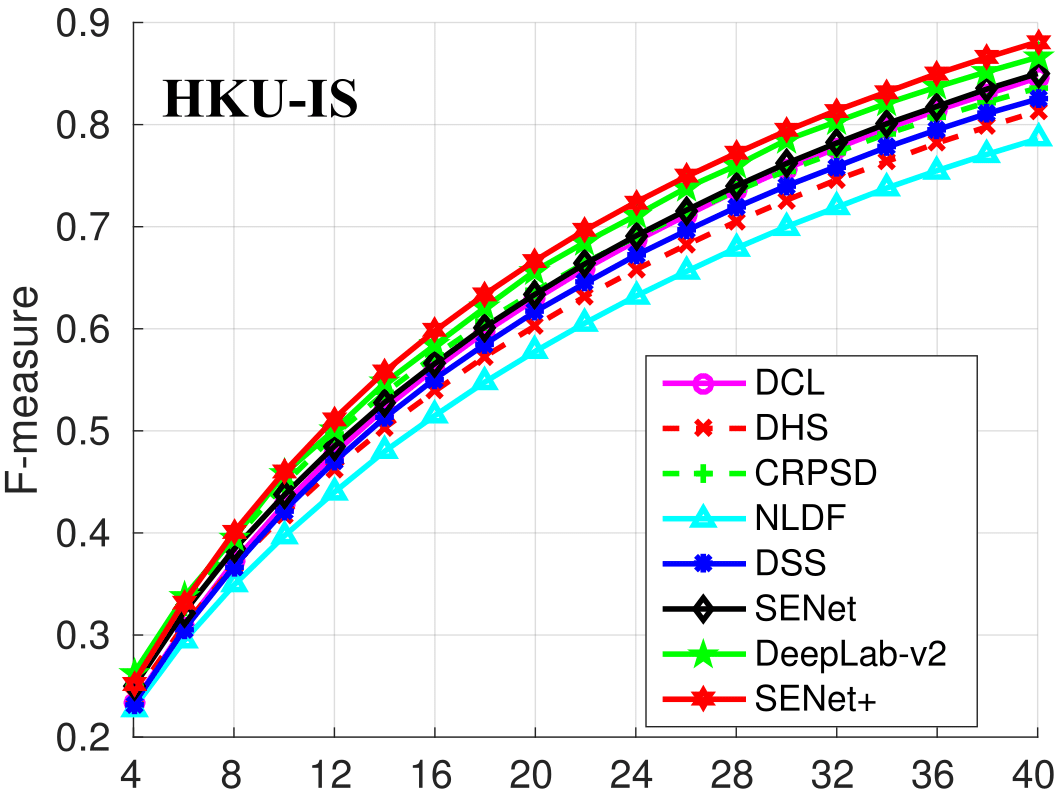}} \\
	\subfigure[PASCAL-S]{\includegraphics[width=4.35cm]{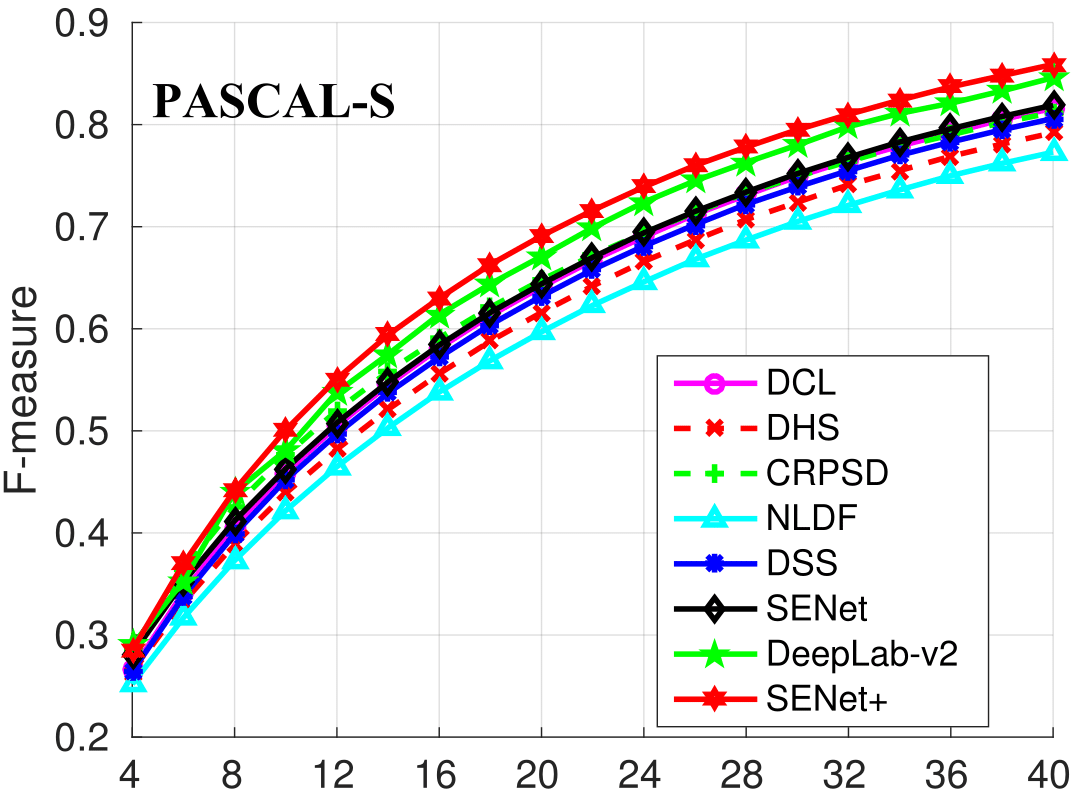}}
	\subfigure[SOD]{\includegraphics[width=4.35cm]{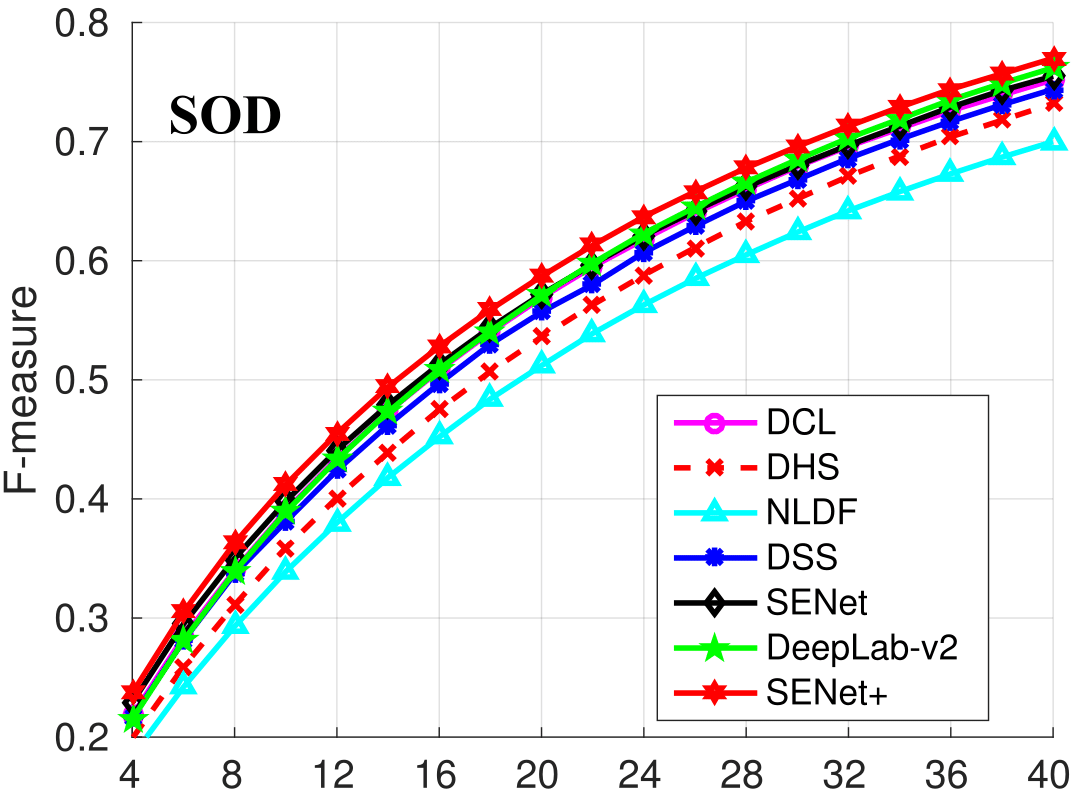}}
	\caption{Comparison of performance along salient object boundaries on 3 datasets. (a) Trimap examples (top-left: image. top-right: ground truth. bottom-left: trimap of 8 pixels. bottom-right: trimap of 20 pixels). (b)-(d) F-measure performance on HKU-IS, PASCAL-S and SOD. To fairly compare the effect along object boundaries, all the results do not go through any post-processing, \eg CRF~\cite{koltun2011efficient}. Best viewed in color. \label{bnd}}
\end{figure}

\begin{figure*}[t]
	\begin{center}
		\subfigure[DHS~\cite{liu2016dhsnet} vs. SENet]{\includegraphics[width=0.32\linewidth]{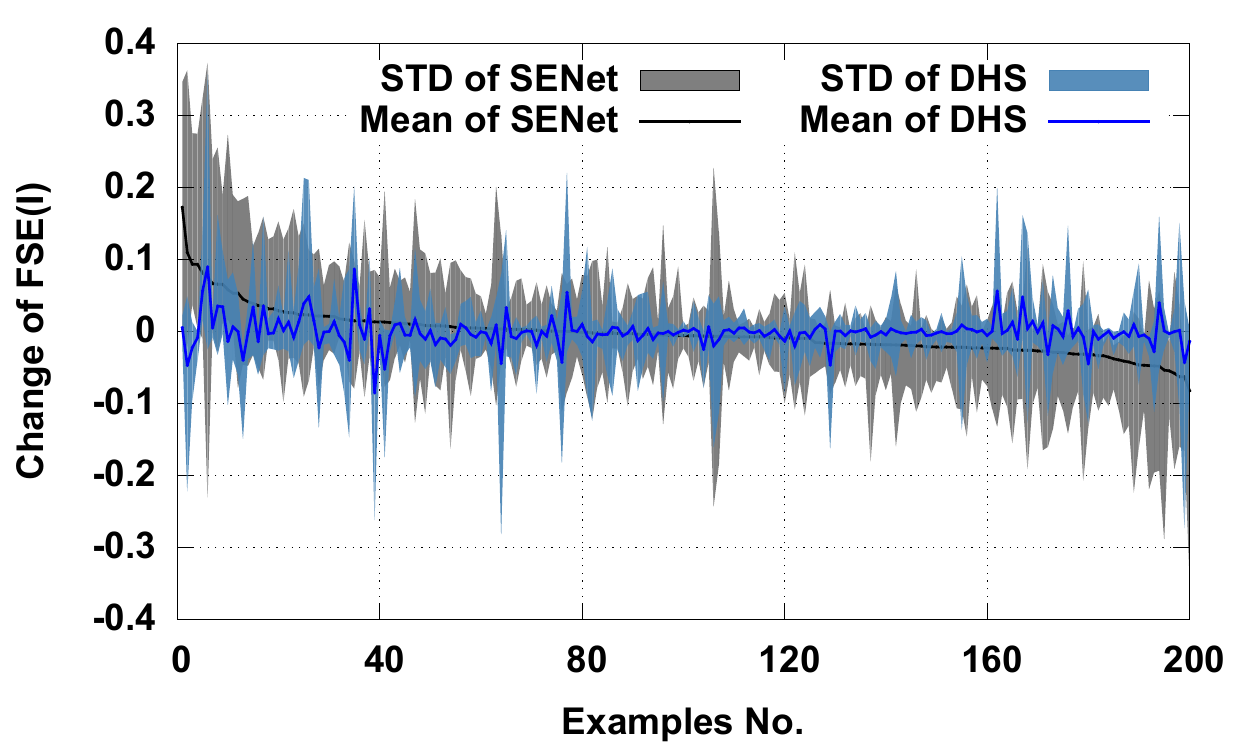}\label{exp_a}} 
		\subfigure[DCL~\cite{li2016deep} vs. SENet]{\includegraphics[width=0.32\linewidth]{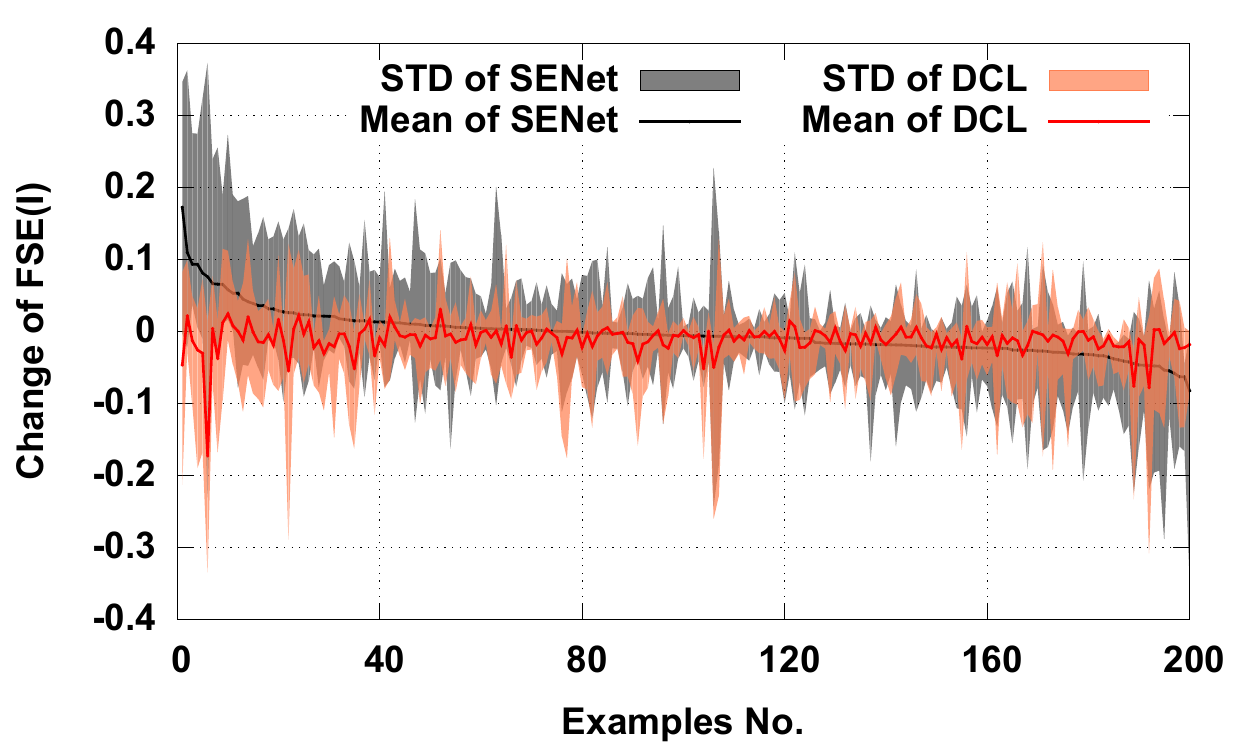}} 
		\subfigure[DSS~\cite{hou2016deeply} vs. SENet]{\includegraphics[width=0.32\linewidth]{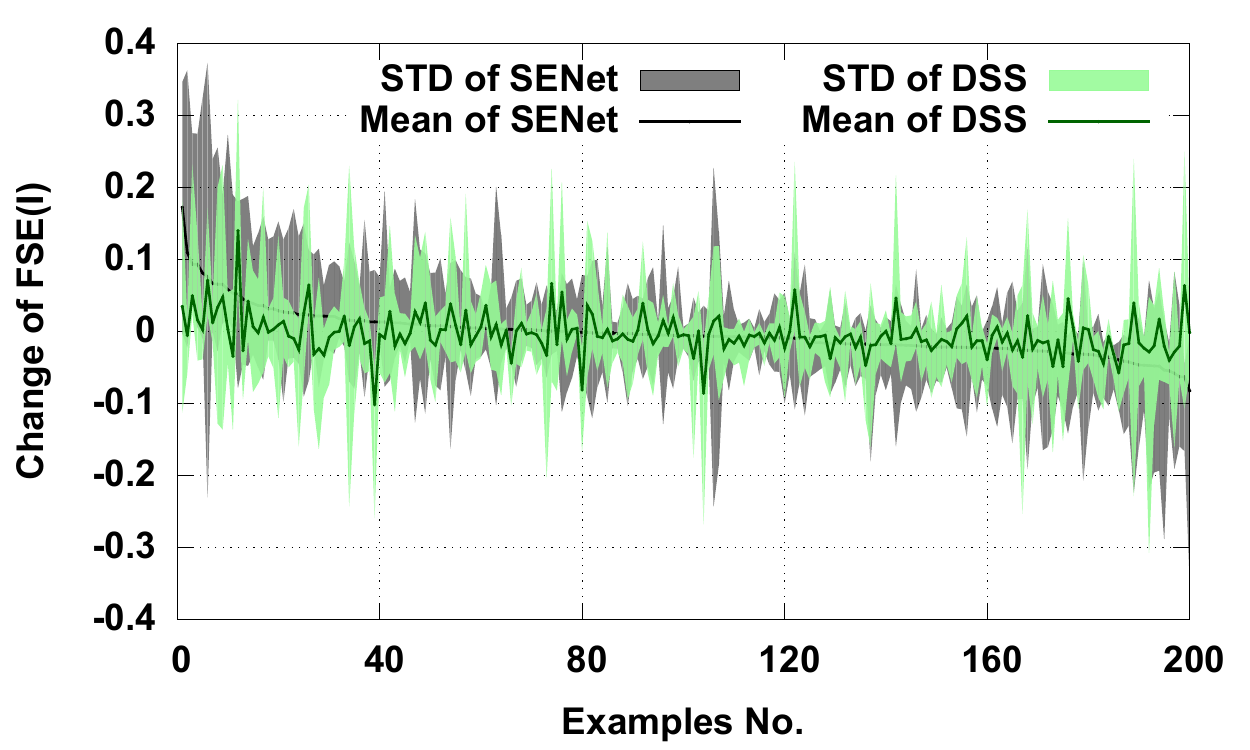}} 
	\end{center}
	\caption{Statistical distribution of ${\rm FSE}(I)$ by different methods. Over the new-built datasets of 200 images from ECSSD, HKU-IS, PASCAL-S and SOD respectively, we calculate the mean and standard deviation (STD) of ${\rm FSE}(I)$ over each examples and plot the distribution in descending order of the mean values of SENet. Three DNN based approaches, \ie, DHS～\cite{liu2016dhsnet}, DCL～\cite{li2016deep} and DSS～\cite{hou2016deeply}, are chosen to compare with the proposed SENet. All methods are evaluated without any post-processing which aims to fairly compare the capability of the saliency detection model. Best viewed in color.}
	\label{exp}
\end{figure*}

In summary, the visual results in Figure~\ref{viscom} illustrate that the proposed SENet can handle various challenges for salient object detection better than  multi-networks architecture~\cite{li2015visual, tang2016saliency} or multi-stages refinement~\cite{wang2016saliency, li2016deep}. We attribute this superiority to the dense connections of classifier on low- and high-level feature maps as well as the specifically designed context blocks.

\subsubsection{Comparison on object boundary detection} To demonstrate the proposed SENet effectively exploits features from low-level layers to better localize boundaries, we quantitatively evaluate  the performance of different models for  the salient object boundary detection. The  evaluation  setting is  similar to~\cite{chen2016deeplab, koltun2011efficient}. We add a narrow band (called trimap in Figure~\ref{bnd_a}) surrounding actual object boundaries which are inferred from the given ground truth. Then we compute the F-measure for these pixels lying within the narrow band. Based on the results in Table~\ref{tab1}, we select the best-performing DHS~\cite{liu2016dhsnet}, DCL~\cite{li2016deep}, CPRSD~\cite{tang2016saliency}, NLDF~\cite{luo2017non}, DSS~\cite{hou2016deeply} and DeepLab-v2~\cite{chen2016deeplab} as the baselines for comparison. To fairly compare the effect along object boundaries, we do not apply any post-processing over all the results. As shown in Figure~\ref{bnd}, exploiting powerful pre-trained network and dense connections with multi-levels features enables  SENet$^\dagger$ to outperform all compared methods across all the  trimap widths and datasets. The VGG16-based SENet has similar performance as DCL~\cite{li2016deep} and CPRSD~\cite{tang2016saliency} which both benefit from the superpixel-based local network component. The proposed dense structure performs  competitively on localizing the object boundaries without introducing additional network. Such efficiency makes the SENet architecture very appealing in practice. In the next subsection, we will further explain the effectiveness of SENet by visualizing the saliency explanation extracted by the self-explanatory generator. 

\subsection{ Explanation on Saliency Detection}
For a given image $I$, we compute and visualize the saliency explanation ${\rm FSE}(I)$ in Eqn. \eqref{eq8}  to explain the behavior of proposed saliency encoder. We also  compare  capability of different DNN based saliency detection models using such evidence. A testing dataset of 200 images is built by extracting 50 images randomly from ECSSD, HKU-IS, PASCAL-S and SOD respectively. MSRA-B is excluded due to the training set of DHS～\cite{liu2016dhsnet} including the entire MSRA-B. Three DNN based approaches, \ie, DHS～\cite{liu2016dhsnet}, DCL～\cite{li2016deep} and DSS～\cite{hou2016deeply}, are chosen to compare with SENet over the built datasets. All methods are evaluated without any post-processing for fair comparison. The F-measure of DHS～\cite{liu2016dhsnet}, DCL～\cite{li2016deep}, DSS～\cite{hou2016deeply} and SENet over the datasets is 0.842, 0.850, 0.865 and 0.877 respectively.

 Figure~\ref{exp} collects and shows statistics of ${\rm FSE}(I)$ for different approaches. For each given image $I$, we calculate the mean and standard deviation (STD) of ${\rm FSE}(I)$ and plot its distribution in descending order of the mean values from SENet. For clear illustration, we compare SENet with DHS～\cite{liu2016dhsnet}, DCL～\cite{li2016deep} and DSS～\cite{hou2016deeply} respectively in Figure~\ref{exp}. Comparing results shown in the three figures, one can observe that the better the saliency model is (based on the F-measure), the larger  variance of ${\rm FSE}(I)$ the model intends to provide, especially in Figure~\ref{exp_a}. The average STD of positive and negative evidence by SENet is 0.0647 and 0.0681 while these number by DHS～\cite{liu2016dhsnet} is 0.0376 and 0.0410. The gap of saliency detection performance between DSS~～\cite{hou2016deeply} and SENet is smaller than the one of DHS～\cite{liu2016dhsnet}. Hence DSS～\cite{hou2016deeply} also  has a larger average STD of ${\rm FSE}(I)$  (0.0522 and 0.0581) than DHS～\cite{liu2016dhsnet}. The larger variance of ${\rm FSE}(I)$ means the saliency explanation of interesting model is more sensitive. A better saliency model intends to offer more information in its corresponding saliency explanation.

\begin{figure}[t]
	\centering
	\subfigure[Source]{\includegraphics[width=1.6cm]{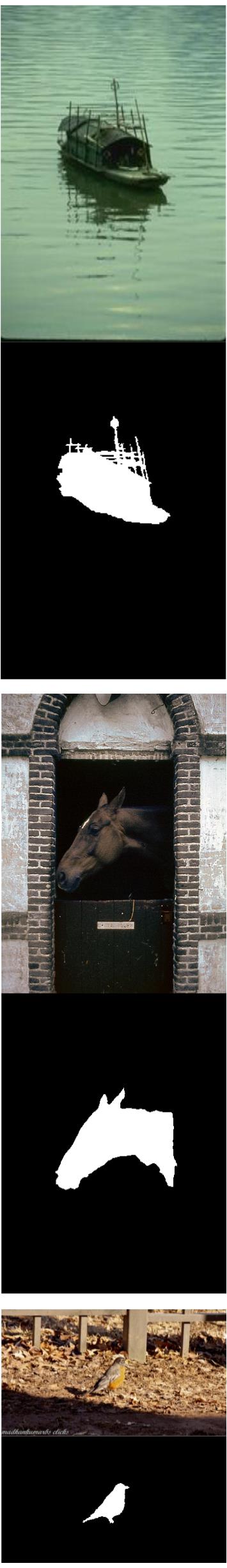}}
	\subfigure[DHS~\cite{liu2016dhsnet}]{\includegraphics[width=1.6cm]{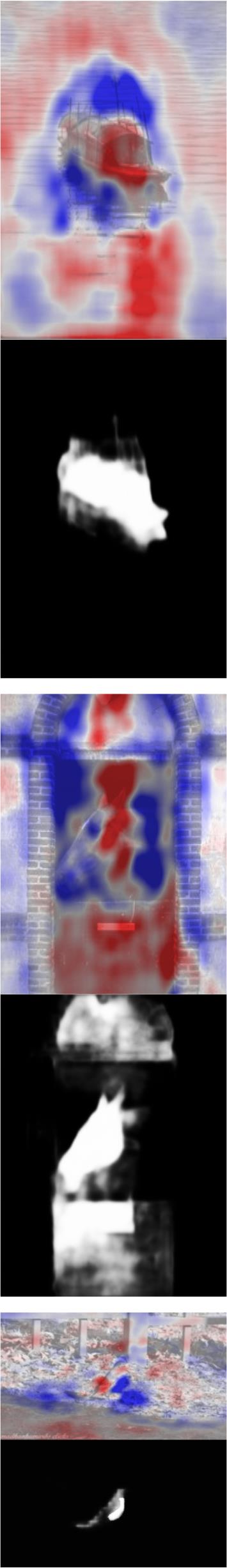}}
	\subfigure[DCL~\cite{li2016deep}]{\includegraphics[width=1.6cm]{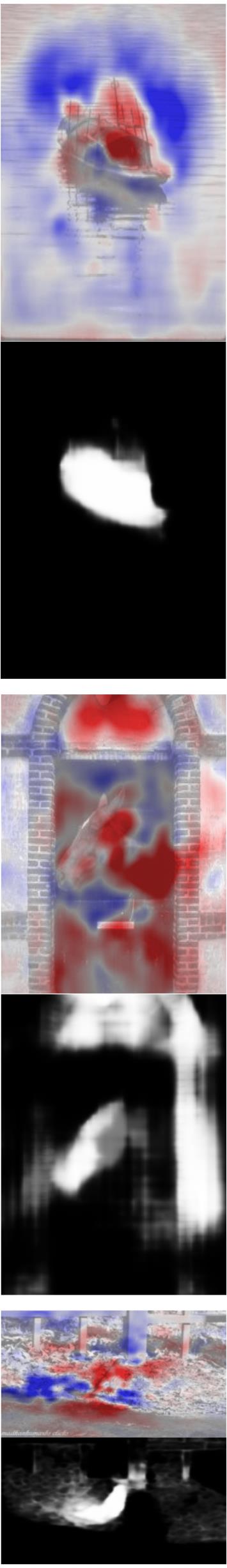}}
	\subfigure[DSS~\cite{hou2016deeply}]{\includegraphics[width=1.6cm]{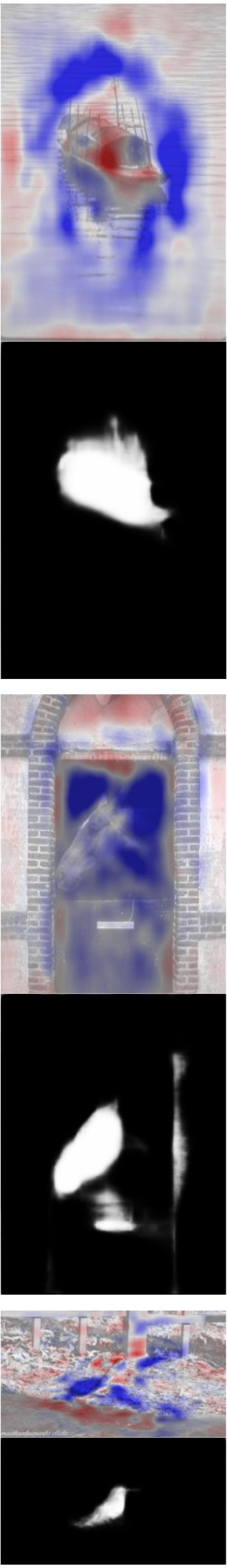}}
	\subfigure[SENet]{\includegraphics[width=1.6cm]{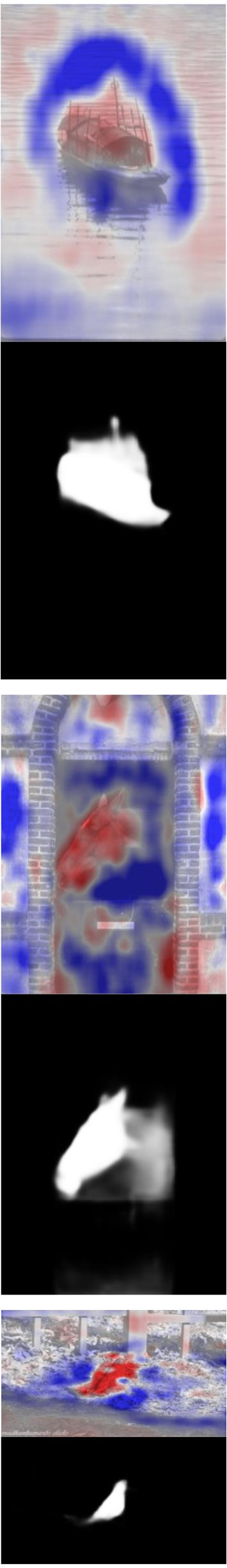}}
	\caption{Visual comparison of saliency results and explanation by different methods. In the first, third and fifth rows, we visualize the saliency explanation that interprets how the saliency prediction made by (b) DHS\cite{liu2016dhsnet}, (c) DCL\cite{li2016deep}, (d) DSS\cite{hou2016deeply},  and (e) proposed SENet. The red areas mean these regions provide positive support for salient object detection while the blue ones represent that the existing of this region could suppress and distract saliency detection. The white color areas mean no significant sensitivity for salient object detection. The second, forth and last rows show the saliency detection results without post-processing. Best viewed in color.   \label{vis_exp}}
\end{figure}
 
 Figure \ref{vis_exp} visualizes the ${\rm FSE}(I)$ of DHS \cite{liu2016dhsnet}, DCL~\cite{li2016deep}, DSS～\cite{hou2016deeply} and the proposed SENet. The red areas indicate  the regions that provide positive support for salient object detection while the blue ones represent  the regions whose presence could suppress and distract saliency detection. The white color areas mean no significant sensitivity for salient object detection. For the example,  in the top tow rows, the \textit{ boat} is easy to detect  for all the models. However, the saliency explanation varies across different models. DSS～\cite{hou2016deeply} and the proposed SENet give heavy negative credits on  surroundings of the boat.In contrast,  DCL~\cite{li2016deep} focuses equally on the salient and non-salient regions. For the second example, DHS~\cite{liu2016dhsnet} and DCL~\cite{li2016deep} regard the \textit{ wall} as positive evidence for saliency detection. This well explains why these two methods detect the wall as the saliency regions in the final saliency maps. In the last example, the \textit{ bird} in the image has similar visual contrast as the background where all the models  label the region around the bird as negative saliency evidence. It indicates that removing this region would help saliency detection. Compared to other three models, the proposed SENet concentrates more on the correct saliency evidence. This leads to more accurate saliency map for SENet. Observing these explanation results in Figure~\ref{fig1} and Figure~\ref{vis_exp}, one can find that a good saliency detection model prefers giving positive emphasis to the context of salient object as useful saliency clues, and regards  surroundings of the salient object as negative factors that restrains the range of the salient regions.

\begin{figure}[t]
	\centering
	\subfigure[]{\includegraphics[width=2cm]{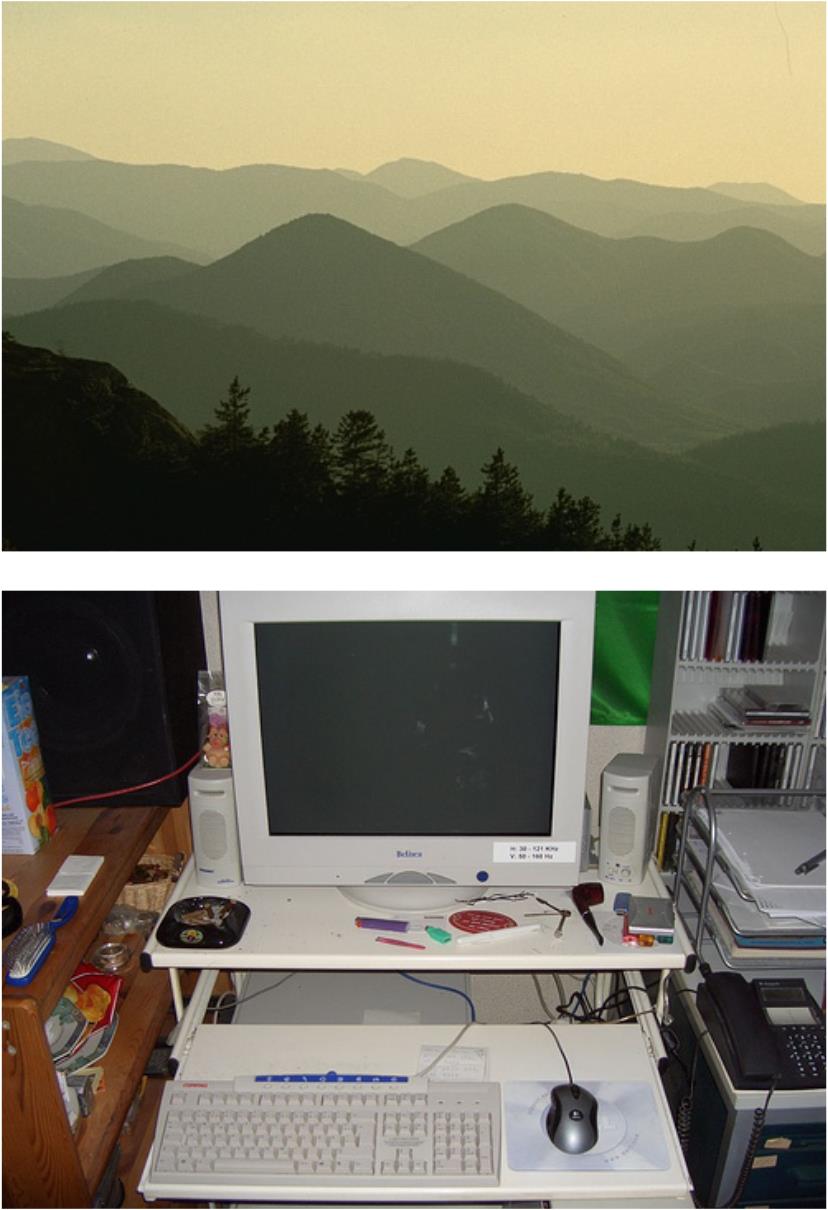}}
	\subfigure[]{\includegraphics[width=2cm]{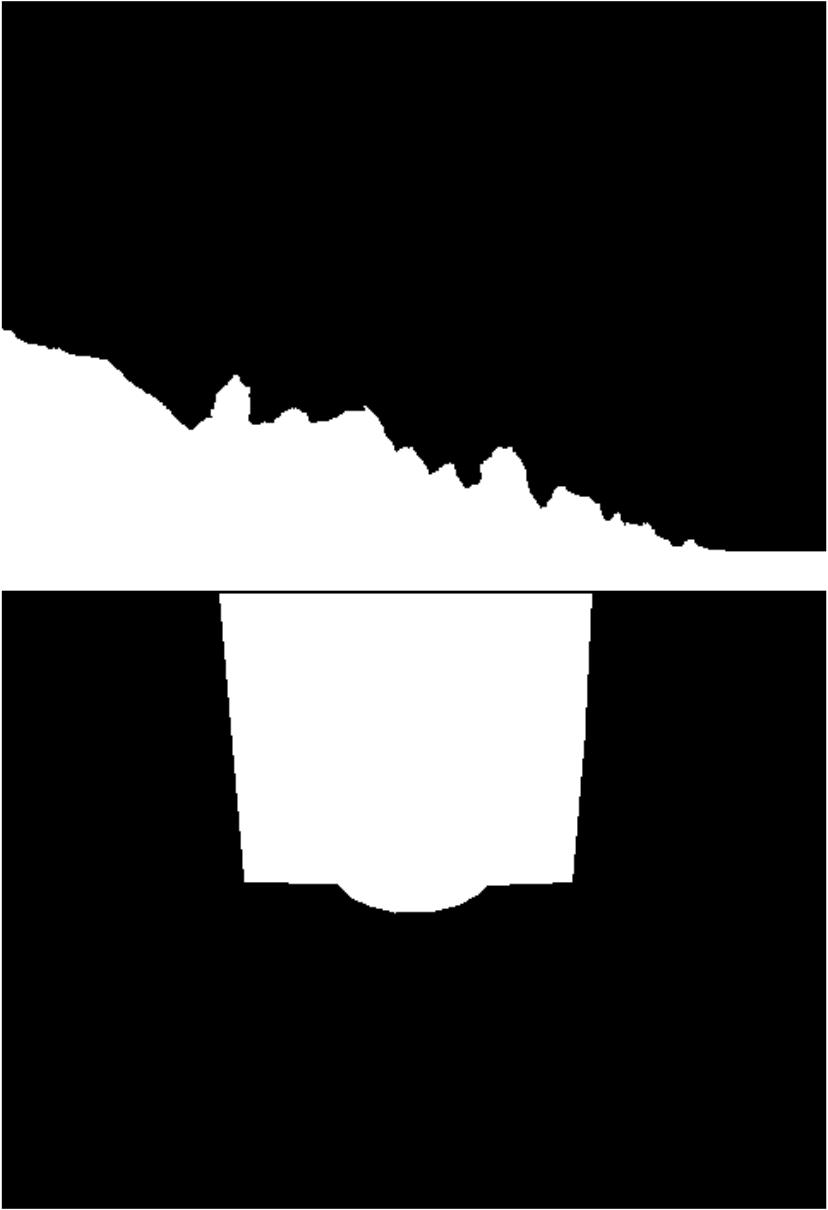}}
	\subfigure[]{\includegraphics[width=2cm]{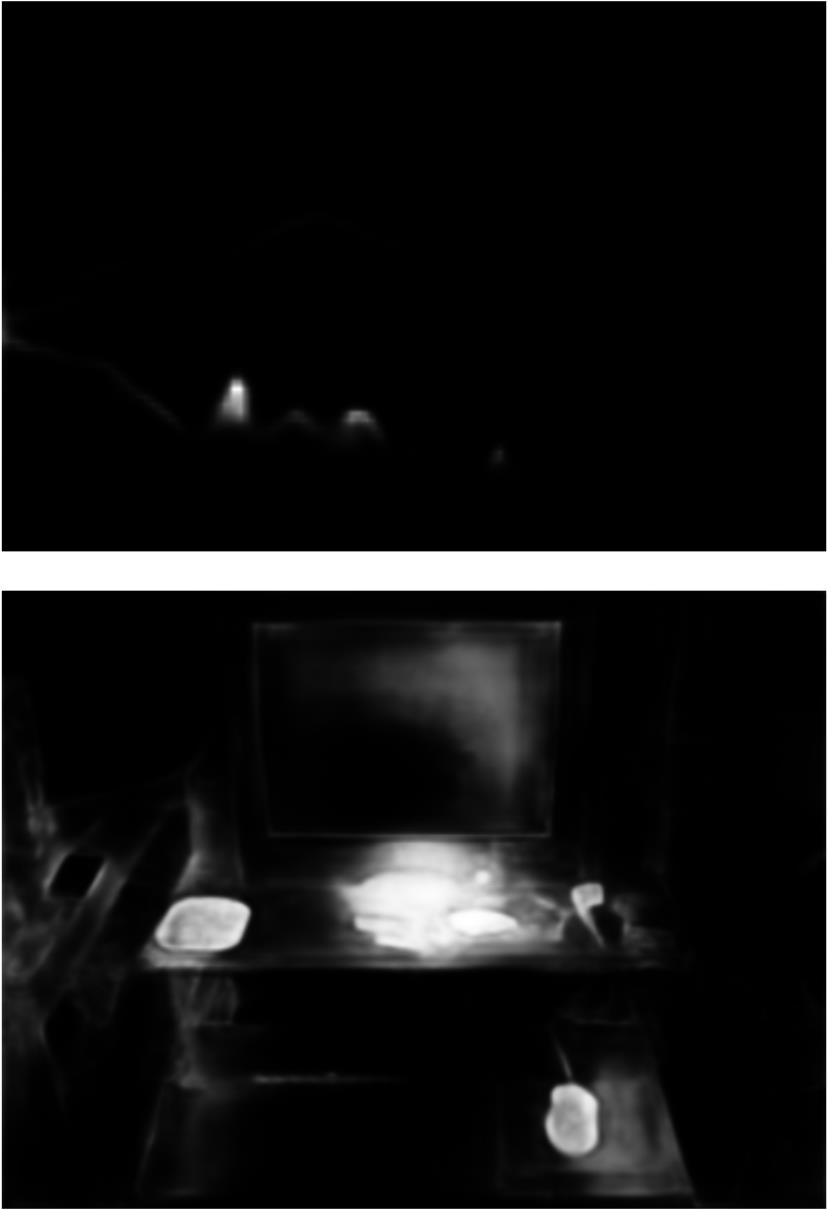}}
	\subfigure[]{\includegraphics[width=2cm]{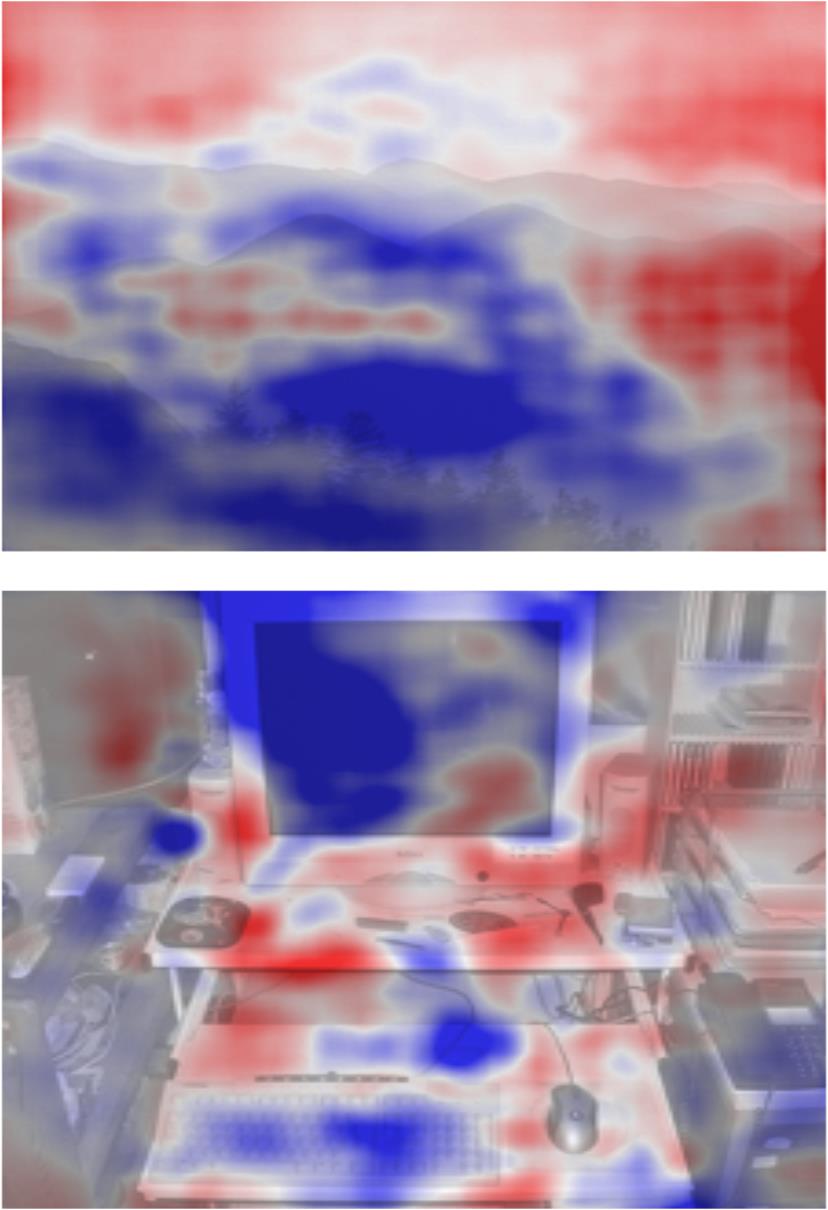}}
	\caption{Some failure cases of SENet. (a) Testing images. (b) Ground truth. (c) Saliency maps. (d) Saliency explanation. Best viewed in color. \label{fig6}}
\end{figure}

Some failure cases of the proposed SENet are shown in Figure \ref{fig6}. We visualize the saliency explanation to interpret why SENet fails to detect these salient objects. For the example, in the first row, the true salient region is regarded as highly negative evidence for saliency detection. SENet regards the meaningless sky as the supportive clues for saliency predictions. Without other distinct object context, only the sky evidence is not enough to detect salient object. In the second example, SENet shows to focus more on the objects that have higher contrast with the background, such as the \textit{ black mouse}. The real saliency object, \ie, the \textit{ monitor}, is regarded as negative for  detection. We attribute such failure to the fact that the objects have more attractive and higher contrast  in the testing image. 

\begin{table}\setlength{\tabcolsep}{2pt}
	\centering
	\caption{Ablation experiments on dataset PASCAL-S \cite{martin2001database}. \label{tab3}}
	\footnotesize
	\begin{tabular}{c|L{5.5cm}|c}
		\toprule
		No. & settings  & $F_\omega$  \\
		\midrule
		\multicolumn{3}{l}{(a) Comparison of short connections:}\\
		1 & baseline$_{S}$ & 0.830 \\
		2 & + {\ttfamily res4b22} & 0.843\\
		3 & + {\ttfamily res4b22} + {\ttfamily res3b3} & 0.849 \\
		4 & + {\ttfamily res4b22} + {\ttfamily res3b3} + {\ttfamily res2c} & 0.852  \\
		5 & + {\ttfamily res4b22} + {\ttfamily res3b3} + {\ttfamily res2c} + {\ttfamily conv1} & 0.853  \\
		\midrule
		\multicolumn{3}{l}{(b) Comparison of long connections:}\\
		6 & baseline$_L$ (No. 4 setting) & 0.852\\
		7 & dense connections only on {\ttfamily res2c} & 0.857 \\
		8 & dense connections on all stages & 0.861\\
		\midrule
		\multicolumn{3}{l}{(c) Comparison of context block:}\\
		9 & baseline$_C$ (No. 4 setting) & 0.852\\
		10 & + context block &0.856 \\
		11 & + context block + No. 8 setting & 0.864\\
		\bottomrule
	\end{tabular}
\end{table}

\subsection{Ablation Study}
To analyze the contributions of each components in the proposed method, we evaluate several variants of SENet$^\dagger$ with various design options. All the ablation experiments are conducted  based on the ResNet-101 \cite{he2016deep} that can be easily trained end-to-end. 

To investigate  the effectiveness of the  dense connections, we separately evaluate  performance of the models with only short or long range dense connections. The baseline for short connections (baseline$_S$ in Table \ref{tab3}) is  a network with architecture similar to   FCN-32s  \cite{long2015fully}. Then we up-sample the feature maps from {\ttfamily res5c} by bilinear interpolation. Before combining with {\ttfamily res4b22} by short connection, a $3\times 3$ convolutional layer is adopted to fine-tune the features. Similarly, different feature maps from {\ttfamily res3b3}, {\ttfamily res2c} and {\ttfamily conv1} are gradually combined what are  denoted as No.3, No.4 and No.5 in Table \ref{tab3}. We  find that  combining with {\ttfamily res4b22} via short connections  can boost the baseline most significantly (bringing 1.34$\%$ improvement). Further combination with {\ttfamily conv1} can only improve the performance incrementally. Therefore, we select the setting of No.4 in Tabel \ref{tab3} as our basic architecture. 

In Table \ref{tab3}(b), we compare two different settings of the long-range connections. For the No.\ 7 setting in Table \ref{tab3}(b), we exploit sparse long connections by concatenating the up-sampled {\ttfamily res5c}, {\ttfamily res4b22} and {\ttfamily res3b3} with {\ttfamily res2c}. This strategy can improve the $F_\omega$ by 0.5$\%$. Such results  confirm that  increasing  connection of classifier to the feature map is beneficial for salient object detection. We further improve the connections by densely concatenating all computed features as in shown in Figure \ref{network}. We observe 0.9$\%$ improvement on PASCAL-S \cite{martin2001database} by exploiting dense connections.

In Table \ref{tab3}(c), we demonstrate  the effectiveness of the context block. We replace the $3\times 3$ convolutional layer after each stage with the proposed context block as shown in Figure \ref{network}. Before the dense connections, all the features are encoded again by the context block. This can improve performance of the baseline$_C$ to 0.858. Overall, the proposed dense connections and context can improve the baseline by 3.4$\%$.

\subsection{Running Time}
It takes  about 20 hours (14 hours) to train the VGG-16 (ResNet-101) based SENet on a single NVIDIA TITAN X GPU and a 3.4 GHz Intel processor. In the phase of inference, it takes about 0.17 second (0.25 second) for the VGG-16 (ResNet-101) based SENet to process an image of size 417$\times$417. Under the same setting to process an image, the running time of state-of-the-art DSS～\cite{hou2016deeply} and DCL~\cite{li2016deep} is about 0.21 second and 1.1 second respectively. Besides, the CRF-based post-processing~\cite{koltun2011efficient} requires extra 0.5 second per image. 

\section{Conclusion} \label{sec5}
We proposed the first Self-Explanatory salient object detection Network (SENet) to provide improved saliency detection method and interpretable evidence. The improved saliency detection method, \ie, the saliency encoder, widens the access of classifier to multi-level clues by reusing multi-level features through introducing dense short- and long-range connections. These dense connections facilitate the encoder to make saliency decisions globally and locally. The self-explanatory generator offers supportive explanation to interpret the saliency decisions by purposely preventing the features of interest from contributing to the saliency encoder. Through visualization, the generator can also compare the capability of different DNN based saliency detection models. Experimental results on five benchmark datasets and the visualized saliency explanation demonstrate the superior performance of the proposed SENet.

%





\ifCLASSOPTIONcaptionsoff
  \newpage
\fi



\bibliographystyle{IEEEtran}
\bibliography{senet}

\end{document}